%% file: main.tex
\documentclass[sigconf]{acmart}
\usepackage{fancyhdr}

\AtBeginDocument{%
  \providecommand\BibTeX{{%
    \normalfont B\kern-0.5em{\scshape i\kern-0.25em b}\kern-0.8em\TeX}}}

\copyrightyear{2020} 
\acmYear{2020} 
\setcopyright{acmcopyright}\acmConference[CCS '20]{Proceedings of the 2020 ACM SIGSAC Conference on Computer and Communications Security}{November 9--13, 2020}{Virtual Event, USA}
\acmBooktitle{Proceedings of the 2020 ACM SIGSAC Conference on Computer and Communications Security (CCS '20), November 9--13, 2020, Virtual Event, USA}
\acmPrice{15.00}
\acmDOI{10.1145/3372297.3417253}
\acmISBN{978-1-4503-7089-9/20/11}

\input{prestuff.tex}

\usepackage{slashbox,balance}
\usepackage{amsthm,amssymb}
\usepackage{diagbox,subcaption}
\usepackage{empheq}
\newenvironment{proof}{\paragraph{Proof:}}{ \hfill$\square$}

\theoremstyle{definition}

\newtheorem{proposition}{Proposition}

\newcommand{\system}{\texttt{IMC}\xspace}

\newcommand*\widefbox[1]{\fbox{\hspace{1em}#1\hspace{1em}}}

\newcommand{\btheta}{\ssub{\theta}{\circ}}

\newcommand{\ayx}{t_{\scaleobj{0.8}{x_{\circ}}}}

\newcommand{\atheta}{\ssub{\theta}{*}}

\newcommand{\calT}{\mathcal{T}}

\newcommand{\lf}{\ell_\mathrm{f}}
\newcommand{\ls}{\ell_\mathrm{s}}

\newcommand{\cifar}{CIFAR10\xspace}
\newcommand{\imgnet}{ImageNet\xspace}
\newcommand{\isic}{ISIC\xspace}
\newcommand{\gtsrb}{GTSRB\xspace}

\newcommand{\bx}{\ssub{x}{\circ}}
\newcommand{\rx}{x}
\newcommand{\rt}{\theta}
\newcommand{\rbx}{\ssub{\bar{x}}{*}}
\newcommand{\rbt}{\ssub{\bar{\theta}}{*}}
\newcommand{\tbt}{\ssub{\tilde{\theta}}{*}}
\newcommand{\rxs}{\ssub{x}{*}}
\newcommand{\rts}{\ssub{\theta}{*}}

\newcommand{\fgsm}{\texttt{FGSM}\xspace}

\newcommand{\pgd}{\texttt{PGD}\xspace}
\newcommand{\cw}{\texttt{C\&W}\xspace}
\newcommand{\bnet}{\texttt{BadNet}\xspace}
\newcommand{\tnet}{\texttt{TrojanNN}\xspace}
\newcommand{\ntnet}{\texttt{TrojanNN}$^*$\xspace}

\newcommand{\nc}{\texttt{NeuralCleanse}\xspace}
\newcommand{\strip}{\texttt{STRIP}\xspace}
\newcommand{\abs}{\texttt{ABS}\xspace}

\newcommand{\ting}[1]{#1}

\newcommand{\another}[1]{#1}

\begin{document}
\fancyhf{}

\title{A Tale of Evil Twins: \\Adversarial Inputs versus Poisoned Models}

\author{Ren Pang}
\email{rbp5354@psu.edu}
\affiliation{%
  \institution{Pennsylvania State University}
}

\author{Hua Shen}
\email{huashen218@psu.edu}
\affiliation{%
  \institution{Pennsylvania State University}
}

\author{Xinyang Zhang}
\email{xqz5366@psu.edu}
\affiliation{%
  \institution{Pennsylvania State University}
}

\author{Shouling Ji}
\email{sji@zju.edu.cn}
\affiliation{%
  \institution{Zhejiang University, Ant Financial}
}

\author{Yevgeniy Vorobeychik}
\email{yvorobeychik@wustl.edu}
\affiliation{%
  \institution{Washington University in St. Louis}
}

\author{Xiapu Luo}
\email{csxluo@comp.polyu.edu.hk}
\affiliation{%
  \institution{Hong Kong Polytechnic University}
}

\author{Alex Liu}
\email{alexliu@antfin.com}
\affiliation{%
  \institution{Ant Financia}
}

\author{Ting Wang}
\email{inbox.ting@gmail.com}
\affiliation{%
  \institution{Pennsylvania State University}
}


\input{abstract.tex}

\ccsdesc[500]{Security and privacy}
\ccsdesc[500]{Computing methodologies~Machine learning}

\keywords{Adversarial attack; Trojaning attack; Backdoor attack}

\maketitle
\thispagestyle{empty}

\input{introduction.tex}

\input{background.tex}
\input{study1.tex}

\input{study2.tex}

\input{study3.tex}

\input{literature.tex}

\input{conclusion.tex}

\begin{acks}
    We thank our shepherd Xiangyu Zhang and anonymous reviewers for valuable feedbacks. 
    This material is based upon work supported by the National Science Foundation under Grant No. 1910546, 1953813, and 1846151. Any opinions, findings, and conclusions or recommendations expressed in this material are those of the author(s) and do not necessarily reflect the views of the National Science Foundation. S. Ji was partly supported by NSFC under No. U1936215, 61772466, and U1836202, the National Key Research and Development Program of China under No. 2018YFB0804102, the Zhejiang Provincial Natural Science Foundation for Distinguished Young Scholars under No. LR19F020003, the Zhejiang Provincial Key R\&D Program under No. 2019C01055, and the Ant Financial Research Funding. X. Luo was partly supported by HK RGC Project (PolyU 152239/18E) and HKPolyU Research Grant (ZVQ8).
\end{acks}

\newpage

\newcommand{\bibpre}{bibs}

\bibliographystyle{ACM-Reference-Format}
\bibliography{\bibpre/aml,\bibpre/debugging,\bibpre/general,\bibpre/ting,\bibpre/optimization}

\input{appendix.tex}

\end{document}

%% file: prestuff.tex
\usepackage{epsfig,amsmath,amsfonts,epsfig,multirow,makecell,caption,soul,csquotes,color,wrapfig,subcaption,mathtools,bm,spverbatim,booktabs,tcolorbox,diagbox,todonotes,amsthm}
\usepackage[e]{esvect}

\def\tablename{Table}
\def\figurename{Figure}

\usepackage{caption}
\makeatletter
\newif\if@restonecol
\makeatother

\usepackage[boxed, ruled, vlined, linesnumbered]{algorithm2e}
\SetKwRepeat{Do}{do}{while}

\newenvironment{changemargin}[2]{\begin{list}{}{
	\setlength{\topsep}{0pt}\setlength{\leftmargin}{0pt}
	\setlength{\rightmargin}{0pt}
	\setlength{\listparindent}{\parindent}
	\setlength{\itemindent}{\parindent}
	\setlength{\parsep}{0pt plus 1pt}
	\addtolength{\leftmargin}{#1}\addtolength{\rightmargin}{#2}
	}\item}
	{\end{list}}

\newenvironment{mitemize}{
	\begin{changemargin}{-3pt}{-0cm}
	\vspace{-10pt}
	\hspace{-5pt}
	\begin{itemize}
	\setlength{\itemsep}{3pt}}
	{\end{itemize}
	\vspace{2pt}
	\end{changemargin}}

\newcommand{\ssup}[2]{{#1}^{\scaleobj{0.8}{#2}}}
\newcommand{\ssub}[2]{{#1}_{\scaleobj{0.8}{#2}}}
\newcommand{\sboth}[3]{{#1}_{\scaleobj{0.8}{#2}}^{\scaleobj{0.8}{#3}}}

\usepackage[first=0,last=9]{lcg}
\usepackage{colortbl}
\definecolor{Gray}{gray}{0.8}
\definecolor{BrickRed}{RGB}{203,65,84}

\usepackage{hyperref}

\newcommand{\msec}[1]{\S\,\ref{#1}}
\newcommand{\mref}[1]{\,\ref{#1}}
\newcommand{\meq}[1]{Eqn\,(\ref{#1})}
\newcommand{\mcite}[1]{\,\cite{#1}}

\usepackage{scalerel}[2016/12/29]

\makeatletter
\providecommand{\leadsfrom}{%
  \mathrel{\mathpalette\reflect@squig\relax}%
}
\newcommand{\reflect@squig}[2]{%
  \reflectbox{$\m@th#1\leadsto$}%
}
\makeatother

\newcommand{\ax}{\ssub{x}{*}}
\newcommand{\ay}{t}

\newcommand{\dnn}{DNN\xspace}
\newcommand{\dnns}{DNNs\xspace}

\def\eqref#1{equation~\ref{#1}}

\def\1{\bm{1}}

\DeclareMathAlphabet{\mathsfit}{\encodingdefault}{\sfdefault}{m}{sl}
\SetMathAlphabet{\mathsfit}{bold}{\encodingdefault}{\sfdefault}{bx}{n}

\def\gD{{\mathcal{D}}}

\def\gF{{\mathcal{F}}}

\def\gR{{\mathcal{R}}}

\def\gT{{\mathcal{T}}}

\def\sE{{\mathbb{E}}}

\def\sR{{\mathbb{R}}}

\DeclareMathOperator{\sign}{sgn}

%% file: abstract.tex
\begin{abstract}

Despite their tremendous success in a range of domains, deep learning systems are inherently susceptible to two types of manipulations: adversarial inputs -- maliciously crafted samples that deceive target deep neural network (DNN) models, and poisoned models -- adversely forged DNNs that misbehave on pre-defined inputs. While prior work has intensively studied the two attack vectors in parallel, there is still a lack of understanding about their fundamental connections: what are the dynamic interactions between the two attack vectors? what are the implications of such interactions for optimizing existing attacks? what are the potential countermeasures against the enhanced attacks? Answering these key questions is crucial for assessing and mitigating the holistic vulnerabilities of DNNs deployed in realistic settings.

Here we take a solid step towards this goal by conducting the first systematic study of the two attack vectors within a unified framework. Specifically, (i) we develop a new attack model that jointly optimizes adversarial inputs and poisoned models; (ii) with both analytical and empirical evidence, we reveal that there exist intriguing ``mutual reinforcement'' effects between the two attack vectors -- leveraging one vector significantly amplifies the effectiveness of the other; (iii) we demonstrate that such effects enable a large design spectrum for the adversary to enhance the existing attacks that exploit both vectors (e.g., backdoor attacks), such as maximizing the attack evasiveness with respect to various detection methods; (iv) finally, we discuss potential countermeasures against such optimized attacks and their technical challenges, pointing to several promising research directions.

\end{abstract}

%% file: introduction.tex
\section{Introduction}
\label{sec:intro}

The abrupt advances in deep learning have led to breakthroughs in a number of long-standing machine learning tasks (e.g., image classification\mcite{Deng:2009:cvpr}, natural language processing\mcite{rajpurkar:squad}, and even playing Go\mcite{silver:nature:alphago}), enabling scenarios previously considered strictly experimental. However, it is now well known that deep learning systems are inherently vulnerable to adversarial manipulations, which significantly hinders their use in security-critical domains, such as autonomous driving, video surveillance, web content filtering, and biometric authentication.

\begin{figure}[!ht]
\begin{center}
  \epsfig{file = 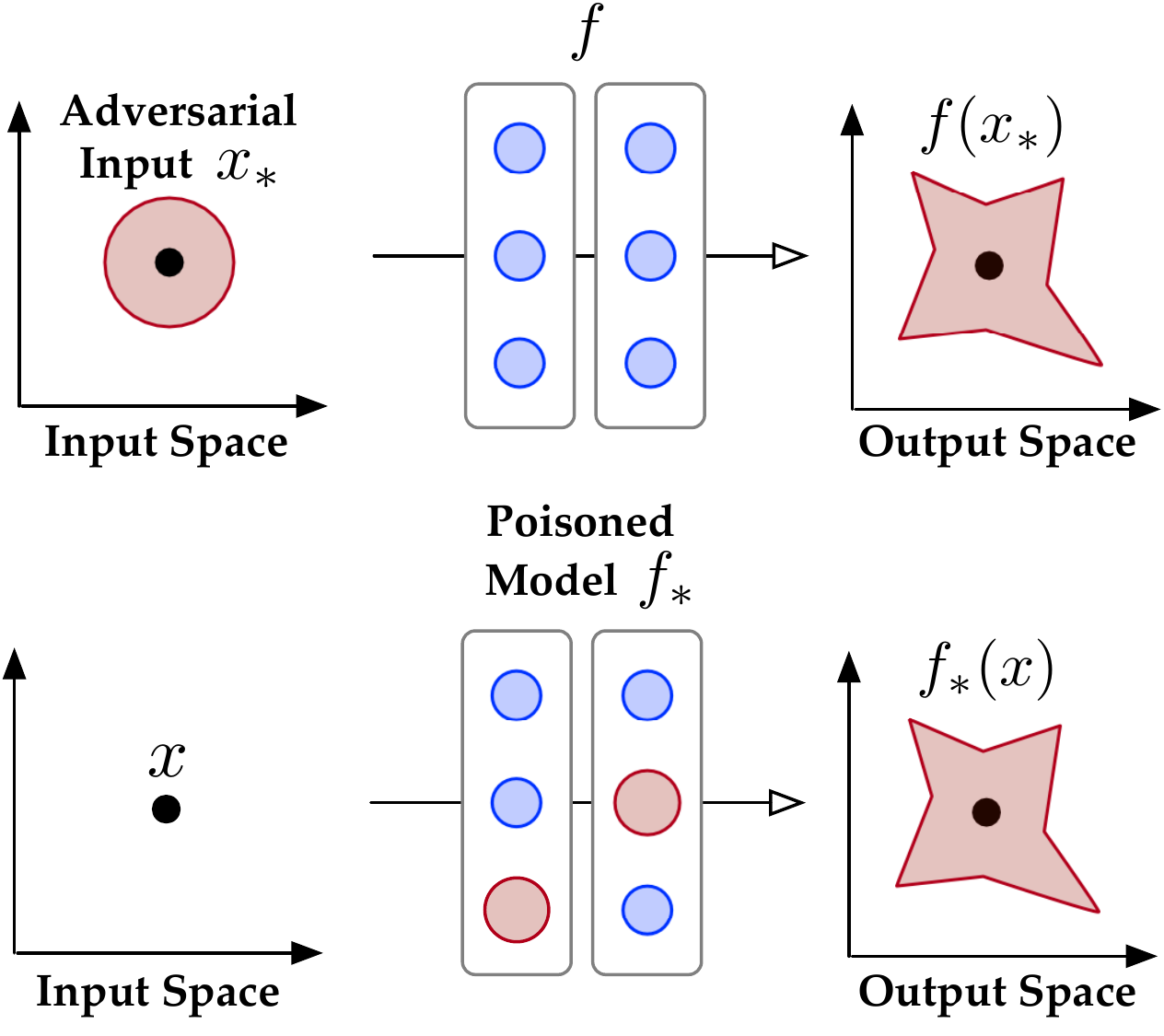, width = 65mm}
\end{center}
\caption{``Duality'' of adversarial inputs and poisoned models. \label{fig:duality}}
\end{figure}

Two primary attack vectors have been considered in the literature. (i) Adversarial inputs -- typically through perturbing a benign input $x$, the adversary crafts an adversarial version $\ax$ which deceives the target \dnn $f$ at inference time\mcite{szegedy:iclr:2014,goodfellow:fsgm,papernot:eurosp:2017,carlini:sp:2017}. (ii) Poisoned models -- during training, the adversary builds malicious functions into $f$, such that the poisoned \dnn $f_*$ misbehaves on one (or more) pre-defined input(s) $x$\mcite{Ji:2017:cns,Suciu:2018:sec,Shafahi:2018:nips,Ji:2018:ccsa}. 
As illustrated in Figure\mref{fig:duality}, the two attack vectors share the same aim of forcing the DNN to misbehave on pre-defined inputs, yet through different routes: one perturbs the input and the other modifies the model. There are attacks (e.g., backdoor attacks\mcite{Gu:arxiv:2017,Liu:2018:ndss}) that leverage the two attack vectors simultaneously: the adversary modifies $f$ to be sensitive to pre-defined trigger patterns (e.g., specific watermarks) during training and  then generates trigger-embedded inputs at inference time to cause the poisoned model $f_*$ to malfunction.

Prior work has intensively studied the two attack vectors separately\mcite{szegedy:iclr:2014,goodfellow:fsgm,papernot:eurosp:2017,carlini:sp:2017,Ji:2017:cns,Suciu:2018:sec,Shafahi:2018:nips,Ji:2018:ccsa}; yet, there is still a lack of understanding about their fundamental connections. First, it remains unclear what the vulnerability to one attack  implies for the other. Revealing such implications is important for developing effective defenses against both attacks. \ting{Further, the adversary may exploit the two vectors together (e.g., backdoor attacks\mcite{Gu:arxiv:2017,Liu:2018:ndss}), or multiple adversaries may collude to perform coordinated attacks.} It is unclear how the two vectors may interact with each other and how their interactions may influence the attack dynamics. Understanding such interactions is critical for building effective defenses against coordinated attacks. Finally, studying the two attack vectors within a unified framework is essential for assessing and mitigating the holistic vulnerabilities of \dnns deployed in practice, in which multiple attacks may be launched simultaneously.

More specifically, in this paper, we seek to answer the following research questions.
\begin{mitemize}
\setlength\itemsep{1.5pt}

\item RQ1 -- {\em What are the fundamental connections between adversarial inputs and poisoned models?}

\item RQ2 -- {\em What are the dynamic interactions between the two attack vectors if they are applied together?}

\item RQ3 -- {\em What are the implications of such interactions for the adversary to optimize the attack strategies?}

\item RQ4 -- {\em What are the potential countermeasures to defend against such enhanced attacks?}

\end{mitemize}

\subsubsection*{\bf Our Work}

This work represents a solid step towards answering the key questions above. We cast adversarial inputs and poisoned models within a unified framework, conduct a systematic study of their interactions, and reveal the implications for \dnns' holistic vulnerabilities, leading to the following interesting findings.

\vspace{2pt}
RA1 -- We develop a new attack model that \ting{jointly optimizes adversarial inputs and poisoned models}. With this framework, we show that there exists an intricate ``duality'' relationship between the two attack vectors. Specifically, they represent different routes to achieve the same aim (i.e., misclassification of the target input): one perturbs the input at the cost of ``fidelity'' (whether the attack retains the original input's perceptual quality), while the other modifies the DNN at the cost of ``specificity'' (whether the attack influences non-target inputs).

\vspace{2pt}
RA2 -- Through empirical studies on benchmark datasets and in security-critical applications (e.g., skin cancer screening\mcite{Esteva:2017:nature}), we reveal that the interactions between the two attack vectors demonstrate intriguing ``mutual-reinforcement'' effects: when launching the unified attack, leveraging one attack vector significantly amplifies the effectiveness of the other (i.e., ``the whole is much greater than the sum of its parts''). We also provide analytical justification for such effects under a simplified setting. 

\vspace{2pt}
RA3 -- Further, \ting{we demonstrate that the mutual reinforcement effects entail a large design spectrum for the adversary to optimize the existing attacks that exploit both attack vectors (e.g., backdoor attacks).} For instance, leveraging such effects, it is possible to enhance the attack evasiveness with respect to multiple defense mechanisms (e.g., adversarial training\mcite{madry:iclr:2018}), which are designed to defend against adversarial inputs or poisoned models alone; it is also possible to enhance the existing backdoor attacks (e.g.,\mcite{Gu:arxiv:2017,Liu:2018:ndss}) with respect to both human vision (in terms of trigger size and transparency) and automated detection methods (in terms of input and model anomaly). 

\vspace{2pt}
RA4 -- Finally, we demonstrate that to effectively defend against such optimized attacks, it is necessary to investigate the attacks from multiple complementary perspectives (i.e., fidelity and specificity) and carefully account for the mutual reinforcement effects in applying the mitigation solutions, which point to a few promising research directions.

\vspace{3pt}
To our best knowledge, this work represents the first systematic study of adversarial inputs and poisoned models within a unified framework. We believe our findings deepen the holistic understanding about the vulnerabilities of \dnns in practical settings and shed light on developing more effective countermeasures.\footnote{\ting{The source code and data  are released at \url{https://github.com/alps-lab/imc}.}}



%% file: background.tex
\section{Preliminaries}
\label{sec:back}

We begin by introducing a set of fundamental concepts and assumptions. Table\mref{tab:symbol} summarizes the important notations in the paper.

\subsection{Deep Neural Networks}

Deep neural networks (\dnns) represent a class of machine learning models to learn high-level abstractions of complex data using multiple processing layers in conjunction with non-linear transformations.
We primarily consider a predictive setting, in which a \dnn $f$ (parameterized by $\theta$) encodes a function $f$: $\mathcal{X} \rightarrow \mathcal{Y}$. Given an input $x \in \mathcal{X}$, $f$ predicts a nominal variable $f(x;\theta)$ ranging over a set of pre-defined classes $\mathcal{Y}$. 

We consider \dnns obtained via supervised learning. To train a model $f$, the training algorithm uses a training set $\mathcal{D}$, of which each instance $(x, y) \in \gD \subset \mathcal{X} \times \mathcal{Y}$ comprises an input $x$ and its ground-truth class $y$. The algorithm determines the best parameter configuration $\theta$ for $f$ via optimizing a loss function $\ell(f(x;\theta), y)$ (e.g., the cross entropy of $y$ and $f(x;\theta)$), which is typically implemented using stochastic gradient descent or its variants\mcite{Zinkevich:2010:nips}.


\subsection{Attack Vectors}

\dnns are inherently susceptible to malicious manipulations. In particular, two primary attack vectors have been considered in the literature, namely, adversarial inputs and poisoned models.

\subsubsection*{\bf Adversarial Inputs}  Adversarial inputs are maliciously crafted samples to deceive target \dnns at inference time. An adversarial input $\ax$ is typically generated by perturbing a benign input $\bx$ to change its classification to a target class $\ay$ desired by the adversary (e.g., pixel perturbation\mcite{madry:iclr:2018} or spatial transformation\mcite{Alaifari:2019:iclr}). 
To ensure the attack evasiveness, the perturbation is often constrained to a {\em feasible set} (e.g., a norm ball $\ssub{\gF}{\epsilon}(\bx) = \{x| \|x - \bx\|_\infty \leq \epsilon\}$). 
Formally, the attack is formulated as the optimization objective:
\begin{align}
  \label{eq:opt2}
   \ax = \arg\min_{x \in \ssub{\gF}{\epsilon}(\bx)  }\,  \ell(x, \ay; \btheta)
\end{align}
where the loss function measures the difference between $f$'s prediction $f(x; \btheta)$ and the adversary's desired classification $\ay$.

\meq{eq:opt2} can be solved in many ways. For instance, {\fgsm}\mcite{goodfellow:fsgm} uses one-step descent along $\ell$'s gradient sign direction, {\pgd}\mcite{madry:iclr:2018} applies a sequence of projected gradient descent steps, while {\cw}\mcite{carlini:sp:2017} solves \meq{eq:opt2} with iterative optimization.


\subsubsection*{\bf Poisoned Models} Poisoned models are adversely forged \dnns that are embedded with malicious functions (i.e., misclassification of target inputs) during training.



This attack can be formulated as perturbing a benign \dnn $\btheta$ to a poisoned version $\atheta$.\footnote{\ting{Note that below we use $\btheta$ ($\atheta$) to denote both a \dnn and its parameter configuration. Also note that the benign model $\btheta$ is independent of the target benign input $\bx$.}}  
To ensure its evasiveness, the perturbation is often constrained to a feasible set $\ssub{\gF}{\delta}(\btheta)$ to limit the impact on non-target inputs. For instance, $\ssub{\gF}{\delta}(\btheta) = \{\theta |  \sE_{x \in \gR} [|f(x;\theta) - f(x;\btheta)|]  \leq \delta  \}$ specifies that the expected difference between $\btheta$ and $\atheta$'s predictions regarding the inputs in a reference set $\gR$ stays below a threshold $\delta$. Formally, the adversary attempts to optimize the  objective function:
\begin{align}
  \label{eq:opt}
\ting{\atheta = \arg\min_{ \theta \in  \ssub{\gF}{\delta}(\btheta) } \sE_{\bx \in \mathcal{T}} \, [\ell(\bx, \ayx; \theta)]}
\end{align}
where \ting{$\mathcal{T}$ represents the set of target inputs, $\ayx$ denotes $\bx$'s classification desired by the adversary, and the loss function is defined similarly as in \meq{eq:opt2}}.

\ting{In practice, \meq{eq:opt} can be solved through either polluting training data\mcite{Gu:arxiv:2017,Suciu:2018:sec,Shafahi:2018:nips} or modifying benign {\dnns}\mcite{Liu:2018:ndss,Ji:2018:ccsa}.} For instance, \texttt{StingRay}\mcite{Suciu:2018:sec} generates poisoning data by perturbing benign inputs close to $\bx$ in the feature space;  \texttt{PoisonFrog}\mcite{Shafahi:2018:nips} synthesizes poisoning data close to $\bx$ in the feature space but perceptually belonging to $\ay$ in the input space; while \texttt{ModelReuse}\mcite{Ji:2018:ccsa} directly perturbs the \dnn parameters to minimize $\bx$'s distance to a representative input from $\ay$ in the feature space.

\subsection{Threat Models}

We assume a threat model wherein the adversary is able to exploit both attack vectors. During training, the adversary forges a \dnn embedded with malicious functions. This poisoned model is then incorporated into the target deep learning system through either system development or maintenance\mcite{Liu:2018:ndss,Ji:2018:ccsa}. At inference time, the adversary further generates adversarial inputs to trigger the target system to malfunction.

\ting{This threat model is realistic. Due to the increasing model complexity and training cost, it becomes not only tempting but also necessary to reuse pre-trained models\mcite{Gu:arxiv:2017,Liu:2018:ndss,Ji:2018:ccsa}. Besides reputable sources (e.g., Google), most pre-trained \dnns on the market (e.g.,\mcite{modelzoo}) are provided by untrusted third parties. Given the widespread use of deep learning in security-critical domains, adversaries are strongly incentivized to build poisoned models, lure users to reuse them, and trigger malicious functions via adversarial inputs during system use.}
The backdoor attacks\mcite{Liu:2018:ndss,Gu:arxiv:2017,Yao:2019:ccs} are concrete instances of this threat model: the adversary makes \dnn sensitive to certain trigger patterns (e.g., watermarks), so that any trigger-embedded inputs are misclassified at inference. Conceptually, one may regard the trigger as a universal perturbation $r$\ting{\mcite{universal-perturbation}}. To train the poisoned model $\atheta$, the adversary samples inputs $\mathcal{T}$ from the training set $\mathcal{D}$ and enforces the trigger-embedded input $(\bx + r)$ for each $\bx \in \mathcal{T}$ to be misclassified to the target class $\ay$. Formally, the adversary optimizes the objective function:
\begin{align}
  \label{eq:opt3}
  \min_{r \in \ssub{\gF}{\epsilon}, \theta \in \ssub{\gF}{\delta}(\btheta)} \sE_{\bx \in \mathcal{T}}\left[ \ell(\bx + r, \ay; \theta) \right]
\end{align}
where both the trigger and poisoned model need to satisfy the evasiveness constraints. Nonetheless, in the existing backdoor attacks, \meq{eq:opt3} is often solved in an ad hoc manner, resulting in suboptimal triggers and/or poisoned models. For example, {\tnet}\mcite{Liu:2018:ndss} pre-defines the trigger shape (e.g., Apple logo) and determines its pixel values in a preprocessing step. We show that the existing attacks can be significantly enhanced within a rigorous optimization framework (details in \msec{sec:study3}).

%% file: study1.tex
\section{A Unified Attack Framework}
\label{sec:study1}

Despite their apparent variations, adversarial inputs and poisoned models share the same objective of forcing target \dnns (modified or not) to misclassify pre-defined inputs (perturbed or not). While intensive research has been conducted on the two attack vectors in parallel, little is known about their fundamental connections.

\subsection{Attack Objectives}
To bridge this gap, we study the two attack vectors using {\em input model co-optimization} (\system), a unified attack framework. Intuitively, within \system, the adversary is allowed to perturb each target input $\bx \in \calT$ and/or to poison the original \dnn $\btheta$, with the objective of forcing the adversarial version 
$\ax$ of each $\bx \in \calT$ to be misclassified to a target class $\ayx$ by the poisoned model $\atheta$. 

Formally, we define the unified attack model by integrating the objectives of \meq{eq:opt2}, \meq{eq:opt}, and \meq{eq:opt3}:
\begin{empheq}[box=\widefbox]{align}
  \label{eq:opt4}
  \min_{\theta \in \ssub{\gF}{\delta}(\btheta) } \sE_{\bx \in \calT} \left[ \min_{x \in \ssub{\gF}{\epsilon}(\bx)}\ell \left(x, \ayx; \theta\right) \right]
\end{empheq}
where the different terms define the adversary's multiple desiderata:

\begin{mitemize}
  \setlength\itemsep{2pt}
  \item The loss $\ell$ quantifies the difference of the model prediction and the classification desired by the adversary, which represents the attack {\em efficacy} -- whether the attack successfully forces the \dnn to misclassify each input $\bx \in \gT$ to its target class $\ayx$.
  \item The constraint $\ssub{\gF}{\epsilon}$ bounds the impact of input perturbation on each target input, which represents the attack {\em fidelity} -- whether the attack retains the perceptual similarity of each adversarial input to its benign counterpart.
  \item The constraint $\ssub{\gF}{\delta}$ bounds the influence of model perturbation on non-target inputs, which represents the attack {\em specificity} -- whether the attack precisely directs its influence to the set of target inputs $\calT$ only.
\end{mitemize}

\begin{figure}[!ht]
  \begin{center}
    \epsfig{file = 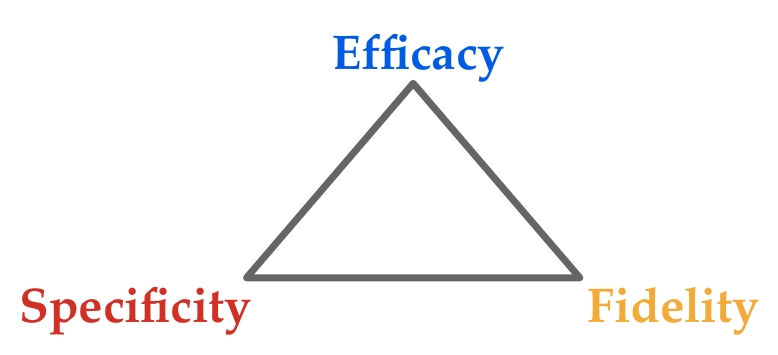, width = 50mm}
  \end{center}
  \caption{Adversary's multiple objectives. \label{fig:triangle}}
\end{figure}

This formulation subsumes many attacks in the literature. Specifically, (i) in the case of $\delta = 0$ and $|\calT| = 1$, \meq{eq:opt4} is instantiated as the adversarial attack; (ii) in the case of $\epsilon = 0$, \meq{eq:opt4} is instantiated as the poisoning attack, which can be either a single target $(|\calT| = 1)$ or multiple targets $(|\calT| > 1)$; and (iii) in the case that a universal perturbation \ting{$(x - \bx)$} is defined for all the inputs \ting{$\{\bx \in \calT\}$} and all the target classes $\{\ayx\}$ are fixed as $\ay$, \meq{eq:opt4} is instantiated as the backdoor attack. Also note that this formulation does not make any assumptions regarding the adversary's capability or resource (e.g.,  access to the training or inference data), while it is solely defined in terms of the adversary's objectives.

Interestingly, the three objectives are tightly intertwined, forming a triangle structure, as illustrated in Figure\mref{fig:triangle}. We have the following observations.
\begin{mitemize}
  \setlength\itemsep{2pt}
  \item It is impossible to achieve all the objectives simultaneously. To attain attack efficacy (i.e., launching a successful attack), it requires either perturbing the input (i.e., at the cost of fidelity) or modifying the model (i.e., at the cost of specificity).
  \item It is feasible to attain two out of the three objectives at the same time. For instance, it is trivial to achieve both attack efficacy and fidelity by setting $\epsilon = 0$ (i.e., only model perturbation is allowed).
  \item With one objective fixed, it is possible to balance the other two. For instance, with fixed attack efficacy, it allows to trade between attack fidelity and specificity.
\end{mitemize}

Next, by casting the attack vectors of adversarial inputs and poisoned models within the \system framework, we reveal their inherent connections and explore the dynamic interactions among the attack efficacy, fidelity, and specificity.

\subsection{Attack Implementation}
\label{sec:impl}

Recall that \system is formulated in \meq{eq:opt4} as optimizing the objectives over both the input and model. While it is impractical to exactly solve \meq{eq:opt4} due to its non-convexity and non-linearity, we reformulate \meq{eq:opt4} to make it amenable for optimization. To ease the discussion, in the following, we assume the case of a single target input $\bx$ in the target set $\gT$ (i.e., $|\calT| = 1$), while the generalization to multiple targets is straightforward. \ting{Further, when the context is clear,  we omit the reference input $\bx$, benign model $\btheta$, and target class $\ay$ to simplify the notations.} 

\subsubsection{\bf Reformulation}

\ting{The constraints $\ssub{\gF}{\epsilon}(\bx)$ and $\ssub{\gF}{\delta}(\btheta)$ in \meq{eq:opt4} essentially bound the fidelity and specificity losses.}
The fidelity loss $\lf(x)$ quantifies whether the perturbed input $x$ faithfully retains its perceptual similarity to its benign counterpart $\bx$ (e.g., $\|x - \bx\|$); the specificity loss $\ls(\theta)$ quantifies whether the attack impacts non-target inputs (e.g., $\sE_{x \in \gR} \left[|f(x;\theta) - f(x;\btheta)|\right]$). According to optimization theory\mcite{convex-optimization}, specifying the bounds $\epsilon$ and $\delta$ on the input and model perturbation amounts to specifying the hyper-parameters $\lambda$ and $\nu$ on the fidelity and specificity losses \ting{(the adversary is able to balance different objectives by controlling $\lambda$ and $\nu$).}
\meq{eq:opt4} can therefore be re-formulated as follows:
\begin{empheq}[box=\widefbox]{align}
  \label{eq:opt5}
  \min_{x, \theta} \ell(x; \theta) + \lambda \lf(x) + \nu \ls(\theta)
\end{empheq}

Nonetheless, it is still difficult to directly optimize \meq{eq:opt5} given that the input $x$ and the model $\theta$ are mutually dependent on each other. \ting{Note that however $\lf$ does not depend on $\theta$ while $\ls$ does not depend on $x$.} We thus further approximate \meq{eq:opt5} with the following bi-optimization formulation: 
\begin{equation}
  \label{eq:opt6}
\left\{
\begin{array}{l}  \ax = \arg\min_x \ell(x; \atheta) + \lambda \lf(x) \\
 \atheta =  \arg\min_{\theta} \ell(\ax; \theta) + \nu \ls(\theta) 
\end{array}
 \right.
\end{equation}

\subsubsection{\bf Optimization}

\begin{figure}[ht!]
  \centering
  \epsfig{file = 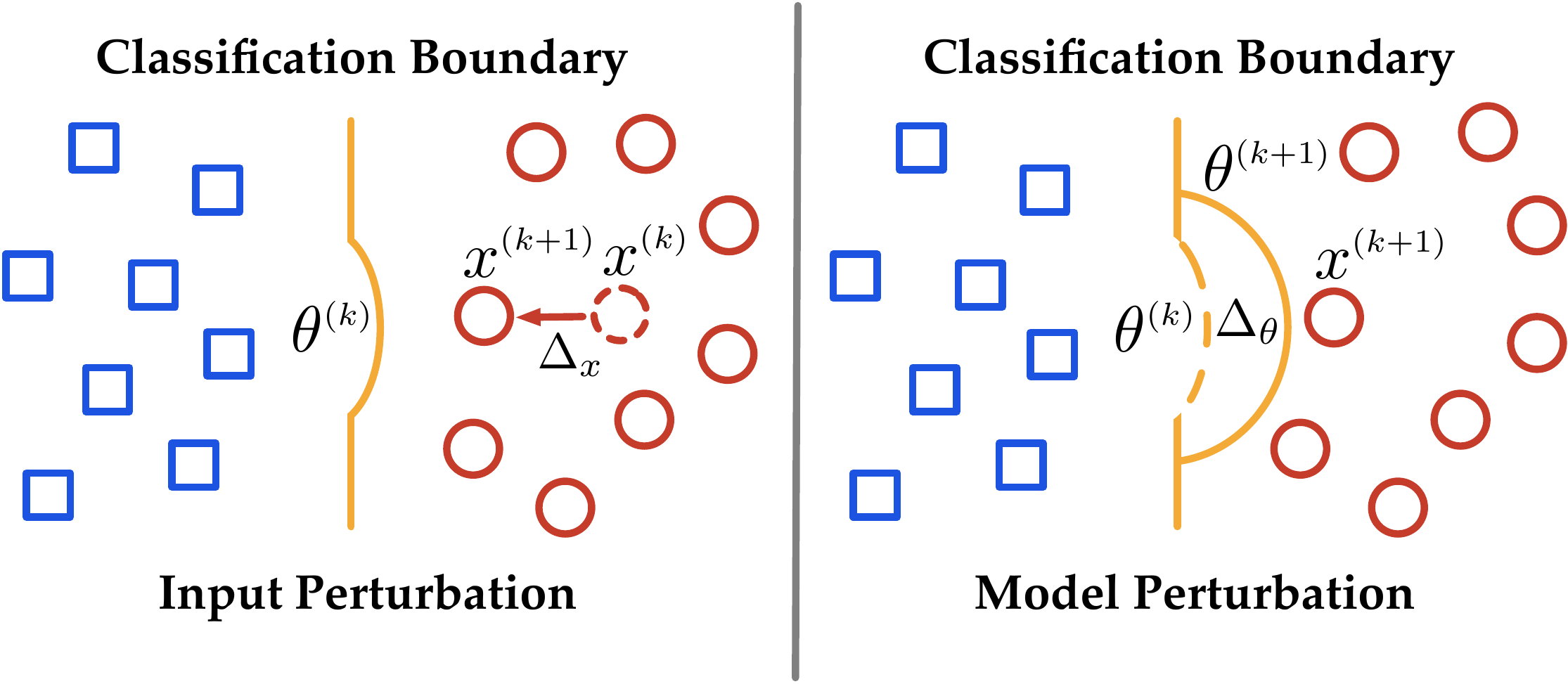, width = 70mm}
  \caption{\system alternates between two operations: (i) input perturbation to update the adversarial input $x$, and (ii) model perturbation to update the poisoned model $\theta$. \label{fig:backdoor}}
\end{figure}

This formulation naturally leads to an optimization procedure that alternates between updating the input $x$ and updating the model $\theta$, as illustrated in Figure\mref{fig:backdoor}. Specifically, let $\ssup{x}{(k)}$ and $\ssup{\theta}{(k)}$ be the perturbed input and model respectively after the $k$-th iteration. The $(k+1)$-th iteration comprises two operations.

\vspace{3pt}
{\em Input Perturbation --} In this step, with the model $\ssup{\theta}{(k)}$ fixed, it updates the perturbed input by optimizing the objective:
\begin{align}
  \ssup{x}{(k+1)}  = \arg\min_{x}\; \ell (x; \ssup{\theta}{(k)})  +  \lambda \lf(x)
\end{align}

\ting{In practice, this step can be approximated by applying an off-the-shelf optimizer (e.g., Adam\mcite{kingma:iclr:2015}) or solved partially by applying gradient descent on the objective function.} For instance, in our implementation, we apply projected gradient descent ({\pgd}\mcite{madry:iclr:2018}) as the update operation:
\begin{displaymath}
  \ssup{x}{(k+1)} = \Pi_{\ssub{\gF}{\epsilon}(\bx)} \left(\ssup{x}{(k)}- \alpha \sign \left(\ssub{\nabla}{x} \ell \left(\ssup{x}{(k)}; \ssup{\theta}{(k)} \right)\right)\right)  
  \end{displaymath}
where $\Pi$ is the projection operator, $\ssub{\gF}{\epsilon}(\bx)$ is the feasible set (i.e., $\{x |  \| x - \bx  \| \leq \epsilon\}$), and $\alpha$ is the learning rate.

\vspace{3pt}
{\em Model Perturbation --} In this step, with the input $\ssup{x}{(k+1)}$ fixed, it searches for the model perturbation by optimizing the objective:
\begin{align}
 \ssup{\theta}{(k+1)}  = \arg\min_{\theta}\; \ell (\ssup{x}{(k+1)}; \theta)  +  \nu \ls(\theta)
\end{align}

\ting{In practice, this step can be approximated by running re-training over a training set that mixes the original training data $\gD$ and $m$ copies of the current adversarial input $\ssup{x}{(k+1)}$.} In our implementation, $m$ is set to be half of the batch size.

\begin{algorithm}[!ht]{\small
  \KwIn{benign input -- $\bx$; benign model -- $\btheta$; target class -- $\ay$; hyper-parameters -- $\lambda, \nu$}
  \KwOut{adversarial input -- $\ax$; poisoned model -- $\atheta$}
  \tcp{\footnotesize initialization}
  $\ssup{x}{(0)}, \ssup{\theta}{(0)}, k \leftarrow \bx, \btheta, 0$\;
  \tcp{\footnotesize optimization}
  \While{not converged yet}{
    \tcp{\footnotesize input perturbation}
    $\ssup{x}{(k+1)}  = \arg\min_{x} \ell (x; \ssup{\theta}{(k)})  +  \lambda \lf(x)$\;
    \tcp{\footnotesize model perturbation}
    $\ssup{\theta}{(k+1)}  = \arg\min_{\theta} \ell (\ssup{x}{(k+1)}; \theta)  +  \nu \ls(\theta)$\;
    $k \leftarrow k + 1$\;
  }
  \Return $(\ssup{x}{(k)}, \ssup{\theta}{(k)})$\;
  \caption{\system Attack \label{alg:attack}}}
\end{algorithm}

\vspace{2pt}
Algorithm\mref{alg:attack} sketches the complete procedure. By alternating between input and model perturbation, it finds \ting{approximately} optimal adversarial input $\ax$ and poisoned model $\atheta$. Note that 
\ting{designed to study the interactions of adversarial inputs and poisoned models (\msec{sec:study2}), Algorithm\mref{alg:attack} is only one possible implementation of \meq{eq:opt4} under the setting of a single target input and both input and model perturbation. To implement other attack variants, one can adjust Algorithm 1 accordingly (\msec{sec:study3}).}
Also note that it is possible to perform multiple input (or model) updates per model (or input) update to accommodate their different convergence rates.

\subsubsection{\bf Analysis}

Next we provide analytical justification for Algorithm\mref{alg:attack}. As \meq{eq:opt5} is effectively equivalent to \meq{eq:opt4}, Algorithm\mref{alg:attack} \ting{approximately} solves \meq{eq:opt5} by alternating between (i) input perturbation -- searching for
$\ax = \arg\min_{x \in \ssub{\gF}{\epsilon}(\bx)} \ell(x;\atheta)$ and (ii) model perturbation -- searching for $\atheta = \arg\min_{\theta \in \ssub{\gF}{\delta}(\btheta)}\ell(\ax;\theta)$. We now show that this implementation effectively solves \meq{eq:opt5} (proof deferred to Appendix A).
\begin{proposition}
  \label{the:danskin}
  Let $\ax \in \ssub{\gF}{\epsilon}(\bx)$ be a minimizer of the function $\min_{x} \ell(x; \theta)$. If $\ax$ is non-zero, then $\nabla_{\theta} \ell(\ax; \theta)$ is a proper descent direction for the objective function of $\min_{x \in \ssub{\gF}{\epsilon}(\bx)} \ell(x; \theta)$.
\end{proposition}
Thus, we can conclude that Algorithm\mref{alg:attack} is an effective implementation of the \system attack framework. It is observed in our empirical evaluation that Algorithm\mref{alg:attack} typically converges within less than 20 iterations (details in \msec{sec:study2}).

%% file: study2.tex
\section{Mutual Reinforcement Effects}
\label{sec:study2}

Next we study the dynamic interactions between adversarial inputs and poisoned models.
With both empirical and analytical evidence, we reveal that there exist intricate ``mutual reinforcement'' effects between the two attack vectors: (i) leverage effect -- with fixed attack efficacy, at slight cost of one metric (i.e., fidelity or specificity), one can disproportionally improve the other metric; (ii) amplification effect -- with one metric fixed, at minimal cost of the other, one can greatly boost the attack efficacy.

\subsection{Study Setting}
\label{sec:setting}


\subsubsection*{\bf Datasets} To factor out the influence of specific datasets, we primarily use 4 benchmark datasets:
\begin{mitemize}
  \setlength\itemsep{2pt}
  \item {\cifar}\mcite{cifar} -- It consists of $32\times32$ color images drawn from 10 classes (e.g., `airplane');
  \item Mini-{\imgnet} -- It is a subset of the ImageNet dataset\mcite{Deng:2009:cvpr}, which consists of $224\times 224$ (center-cropped) color images drawn from 20 classes (e.g., `dog');
  \item {\isic}\mcite{Esteva:2017:nature} -- It represents the skin cancer screening task from the \isic 2018 challenge, in which given $600\times 450$ skin lesion images are categorized into a 7-disease taxonomy (e.g., `melanoma');
  \item {\gtsrb}\mcite{gtsrb} -- It consists of color images of size ranging from $29\times 30$ to $144\times 48$, each representing one of 43 traffic signs.
\end{mitemize}

Note that among these datasets, \isic and \gtsrb in particular represent security-sensitive tasks (i.e., skin cancer screening\mcite{Esteva:2017:nature} and traffic sign recognition\mcite{Bojarski:2016:arxiv}).

\subsubsection*{\bf DNNs}
We apply ResNet18\mcite{resnet} to \cifar, \gtsrb and \imgnet and ResNet101 to \isic as the reference \dnn models. \ting{Their top-1 accuracy on the testset of each dataset is summarized in Table\mref{tab:accuracy}.} Using two distinct \dnns, we intend to factor out the influence of individual \dnn characteristics (e.g., network capacity).

\begin{table}[ht!]{\small
    \begin{tabular}{c|c|c|c|c}
               & \cifar   & \imgnet  & \isic     & \gtsrb   \\
      \hline
      Model    & ResNet18 & ResNet18 & ResNet101 & ResNet18 \\
      \hline
      Accuracy & 95.23\%  & 94.56\%  & 88.18\%   & 99.12\%
    \end{tabular}
    \caption{Accuracy of benign \dnns on reference datasets.\label{tab:accuracy}}}
\end{table}

\subsubsection*{\bf Attacks}
\ting{Besides the \system attack in \msec{sec:impl}, we also implement two variants of \system (with the same hyper-parameter setting) for comparison: (i) input perturbation only, in which \system is instantiated as the adversarial attack (i.e., {\pgd}\mcite{madry:iclr:2018}), and (ii) model perturbation only, in which \system is instantiated as the poisoning attack. The implementation details are deferred to Appendix B.}

\begin{figure*}[ht!]
  \centering
  \epsfig{file = 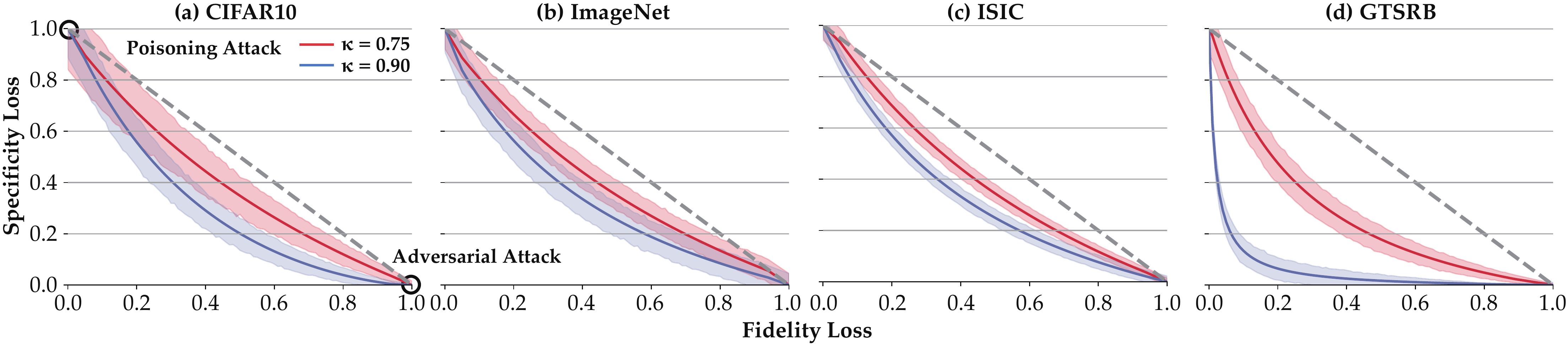, width = 170mm}
  \caption{Disproportionate trade-off between attack fidelity and specificity. \label{fig:fstradeoff}}
\end{figure*}

\subsubsection*{\bf Measures}

We quantify the attack objectives as follows.

\vspace{2pt}
{\em Efficacy --} We measure the attack efficacy by the misclassification confidence, $f_t(\ax; \atheta)$, which is the probability that the adversarial input $\ax$ belongs to the target class $t$ as predicted by the poisoned model $\atheta$. We consider the attack successful if the misclassification confidence exceeds a threshold $\kappa$.

\vspace{2pt}
{\em Fidelity --} We measure the fidelity loss by the $L_p$-norm of the input perturbation $\lf(\ax) \triangleq \|\ax - \bx\|_p$. \ting{Following previous work on adversarial attacks\mcite{goodfellow:fsgm,carlini:sp:2017,madry:iclr:2018}, we use $p = \infty$ by default in the following evaluation.}

\vspace{2pt}
{\em Specificity --} Further, we measure the specificity loss using the difference of the benign and poisoned models on classifying a reference set $\gR$. Let $\mathbb{I}_z$ be the indicator function that returns 1 if $z$ is true and 0 otherwise. The specificity loss can be defined as:
\begin{align}
  \label{eq:sloss}
  \ls(\atheta) \triangleq  \sum_{x \in \gR}\frac{\mathbb{I}_{f(x;\btheta) \neq f(x;\atheta)}}{|\gR|}
\end{align}

With fixed attack efficacy $\kappa$, let $(\ax, \atheta)$ be the adversarial input and poisoned model generated by \system, and $\rbx$ and $\rbt$ be the adversarial input and poisoned model given by the adversarial and poisoning attacks respectively. Because the adversarial and poisoning attacks are special variants of \system, we have $\rxs = \rbx$ if $\rts = \btheta$ and $\rts = \rbt$ if $\rxs = \bx$. Thus, in the following, we normalize the fidelity and specificity losses as $\lf(\rxs)/ \lf(\rbx)$ and $\ls(\rts) / \ls(\rbt)$ respectively, both of which are bound to $[0, 1]$. For reference, the concrete specificity losses $\ls(\rbt)$ (average accuracy drop) caused by the poisoning attack on each dataset are summarized in Table\mref{tab:accuracy-drop}.

\begin{table}[ht!]{\small \ting{
      \begin{tabular}{c|c|c|c|c}
        $\kappa $ & \cifar & \imgnet & \isic  & \gtsrb \\
        \hline
        0.75      & 0.11\% & 2.44\%  & 1.53\% & 0.25\% \\
        \hline
        0.9       & 0.12\% & 3.83\%  & 1.62\% & 0.27\%
      \end{tabular}
      \caption{Specificity losses (average accuracy drop) caused by poisoning attacks on reference datasets.\label{tab:accuracy-drop}}}}
\end{table}


\subsection{Effect I: Leverage Effect}

In the first set of experiments, we show that for fixed attack efficacy, with disproportionally small cost of fidelity, it is feasible to significantly improve the attack specificity, and vice versa.

\subsubsection{\bf Disproportionate Trade-off}

For each dataset, we apply the adversarial, poisoning, and \system attacks against 1,000 inputs randomly sampled from the testset (as the target set $\mathcal{T}$), and use the rest as the reference set $\mathcal{R}$ to measure the specificity loss. For each input of $\mathcal{T}$, we randomly select its target class and fix the required attack efficacy (i.e., misclassification confidence $\kappa$). By varying \system's hyper-parameters $\lambda$ and $\nu$, we control the importance of fidelity and specificity. We then measure the fidelity and specificity losses for all the successful cases. Figure\mref{fig:fstradeoff} illustrates how \system balances fidelity and specificity.
Across all the datasets and models, we have the following observations.

First, with fixed attack efficacy (i.e., $\kappa$ = 0.9), by sacrificing disproportionally small fidelity (i.e., input perturbation magnitude), \system significantly improves the attack specificity (i.e., accuracy drop on non-target inputs), compared with required by the corresponding poisoning attack. For instance, in the case of \gtsrb (Figure\mref{fig:fstradeoff}\,(d)), as the fidelity loss increases from 0 to 0.05, the specificity loss is reduced by more than 0.48.

Second, this effect is symmetric: a slight increase of specificity loss also leads to significant fidelity improvement, compared with required by the corresponding adversarial attack. For instance, in the case of \cifar (Figure\mref{fig:fstradeoff}\,(a)), as the specificity loss increases from 0 to 0.1, the specificity loss drops by 0.37.

Third, higher attack efficacy constraint brings more fidelity-specificity trade-off. Observe that across all datasets, \gtsrb shows significantly larger curvature, which might be explained by the higher model accuracy(99.12\%) and larger number of classes(43).



\begin{tcolorbox}[boxrule=0pt, title= Leverage Effect]
  There exists an intricate fidelity-specificity trade-off. At disproportionally small cost of fidelity, it is possible to significantly improve specificity, and vice versa.
\end{tcolorbox}

\begin{figure*}[ht!]
  \centering
  \epsfig{file = 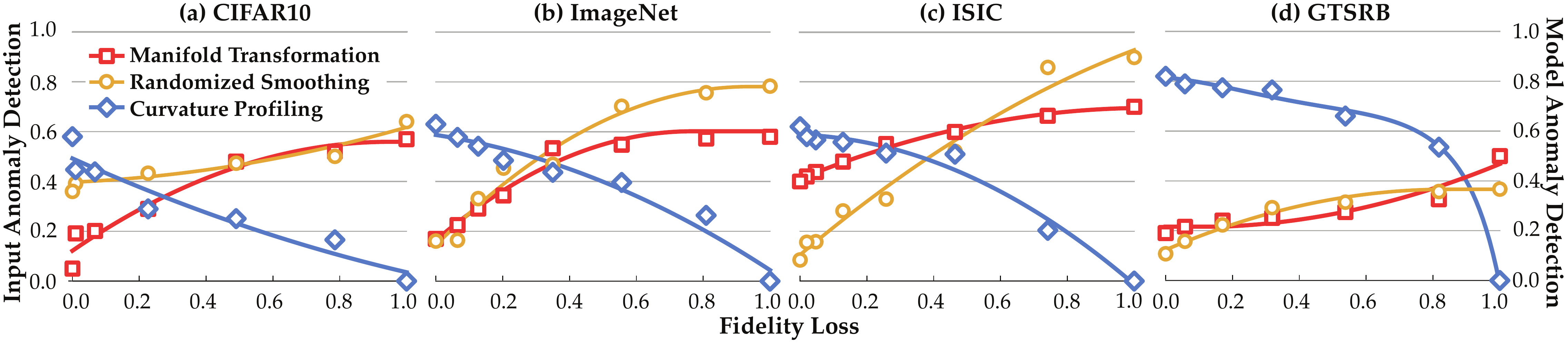, width = 170mm}
  \caption{Detection rates of input anomaly (by manifold projection\mcite{Meng:2017:ccs}) and model anomaly (by curvature profile\mcite{Dezfooli:2018:arxiv}). \label{fig:cross}}
\end{figure*}

\begin{figure*}[ht!]
  \centering
  \epsfig{file = 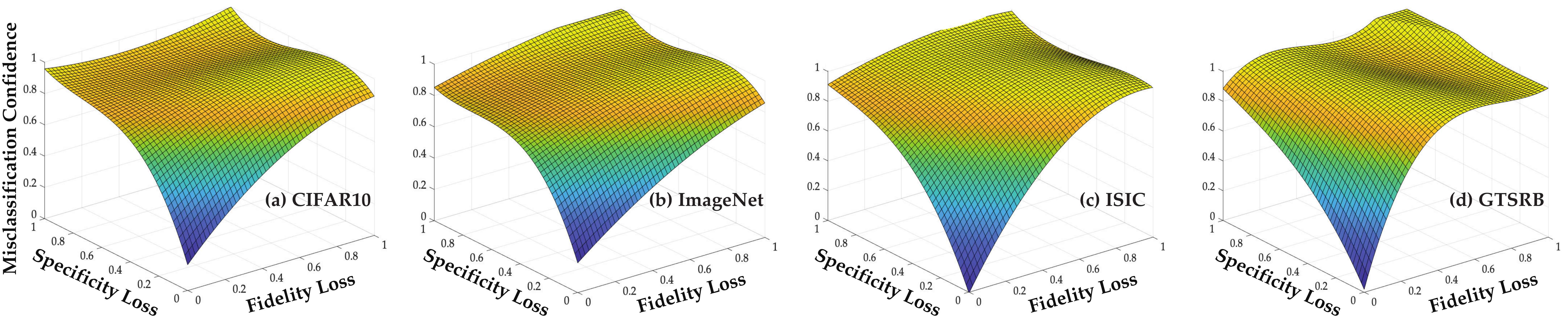, width = 170mm}
  \caption{Average misclassification confidence ($\kappa$) as a function of fidelity and specificity losses. \label{fig:tradeoff3d}}
\end{figure*}

\subsubsection{\bf Empirical Implications}

The leverage effect has profound implications. We show that it entails a large design spectrum for the adversary to optimize the attack evasiveness with respect to various detection methods (detectors). \ting{Note that here we do not consider the adversary's adaptiveness to specific detectors but rather focus on exposing the design spectrum enabled from a detection perspective.} In \msec{sec:study3}, we show that \system also allows to enhance the attacks by adapting to specific detectors. To assess \system's evasiveness, we consider \ting{three} complementary detectors.

\vspace{2pt}
{\em Input Anomaly --} From the input anomaly perspective, we apply manifold transformation\mcite{Meng:2017:ccs} as the detector. At a high level, it employs a reformer network to project given inputs to the manifold spanned by benign inputs and a detector network to differentiate benign and adversarial inputs. \ting{Besides, we apply randomized smoothing\mcite{randomized-smoothing} as another detector, which transforms a given \dnn into a ``smoothed'' model and considers a given input $\ax$ as adversarial if the probability difference of $\ax$'s largest and second largest classes exceeds a threshold.}

\vspace{2pt}
{\em Model Anomaly --} From the model anomaly perspective, we apply curvature profiling\mcite{Dezfooli:2018:arxiv} as the detector. Recall that the poisoning attack twists the classification boundary surrounding the target input $\ax$; thus, the loss function tends to change abruptly in $\ax$'s vicinity. To quantify this property, we compute the eigenvalues of the Hessian $\ssub{H}{x}(\ax) = \sboth{\nabla}{x}{2} \ell(\ax)$. Intuitively, larger (absolute) eigenvalues indicate larger curvatures of the loss function. We define the average (absolute) value of the top-$k$ eigenvalues of $\ssub{H}{x}(\ax)$ as $\ax$'s curvature profile: $\frac{1}{k}\sum_{i=1}^k |\ssub{\lambda}{i} (\ssub{H}{x}(\ax) )|$, where $\ssub{\lambda}{i}(M)$ is the $i$-th eigenvalue of matrix $M$ (details in Appendix B). We compare the curvature profiles of given inputs and benign ones, and use the Kolmogorov–Smirnov statistics to differentiate the two sets.

\vspace{2pt}
We apply the above detectors to the adversarial inputs and poisoned models generated by \system under varying fidelity-specificity trade-off ($\kappa$ fixed as 0.75). Figure\mref{fig:cross} measures the detection rates for different datasets. We have the following observations.

The detection rate of input anomaly grows monotonically with the fidelity loss (i.e., input perturbation magnitude); on the contrary, the detection rate of model anomaly drops quickly with the fidelity loss (i.e., disproportionate specificity improvement due to the leverage effect).
For instance, in the case of \imgnet (Figure\mref{fig:cross}\,(b)), as the fidelity loss varies from 0 to 0.35, the detection rate of input anomaly increases from 0.17 to 0.53 by manifold transformation and from 0.16 to 0.47 by randomized smoothing, while the detection rate of corresponding model anomaly drops from 0.63 to 0.44.

Moreover, across all the cases, \system is able to balance fidelity and specificity, leading to high evasiveness with respect to multiple detectors simultaneously. For instance, in the case of \cifar (Figure\mref{fig:cross}\,(a)), with the fidelity loss set as 0.23, the detection rates of manifold transformation, randomized smoothing, and curvature profiling are reduced to 0.29, 0.43, and 0.29 respectively.

\begin{figure*}[ht!]
  \centering
  \epsfig{file = 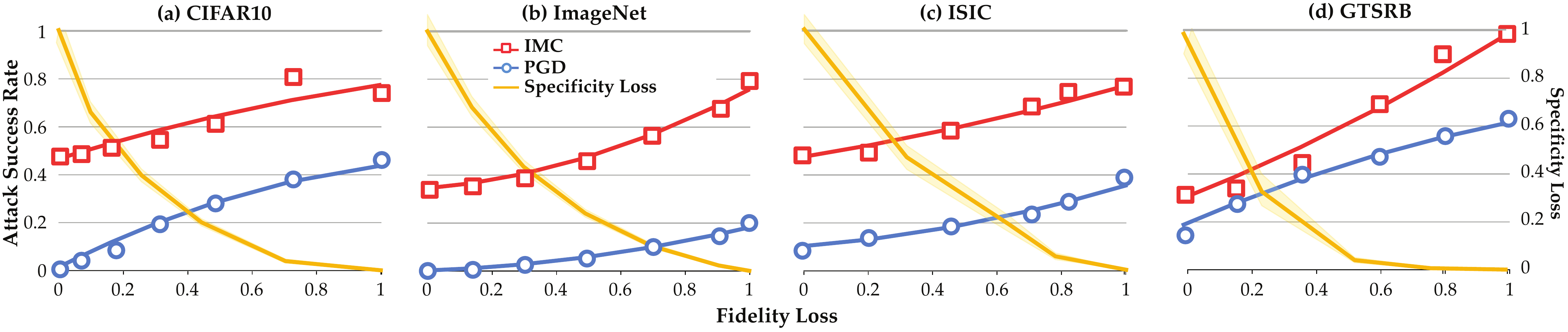, width = 170mm}
  \caption{Accuracy and robustness (with respect to \pgd and \system) of adversarially re-trained models. \label{fig:cross2}}
\end{figure*}

\begin{figure*}[!ht]
  \centering
  \epsfig{file = 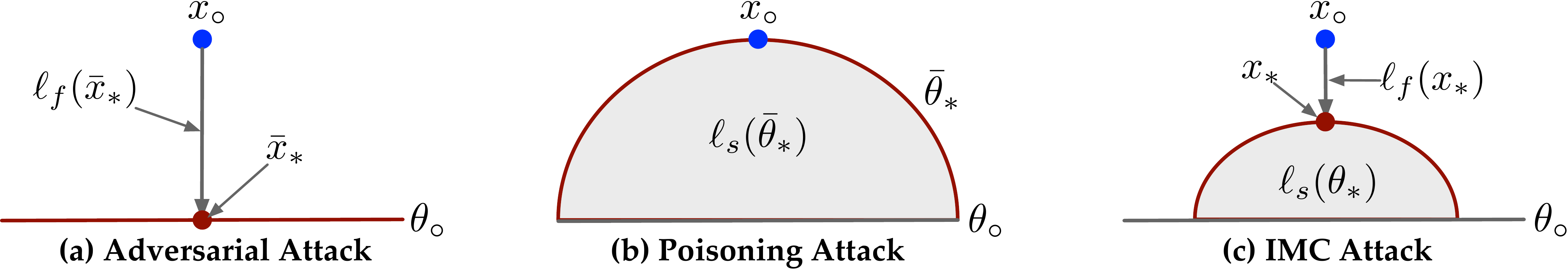, width = 140mm}
  \caption{Comparison of the adversarial, poisoning, and \system attacks under fixed attack efficacy. \label{fig:leverage}}
\end{figure*}

\subsection{Effect II: Amplification Effect}
Next we show that the two attack vectors are able to amplify each other and attain attack efficacy unreachable by each vector alone.

\subsubsection{\bf Mutual Amplification}

We measure the attack efficacy (average misclassification confidence) attainable by the adversarial, poisoning, and \system attacks under varying fidelity and specificity losses. The results are shown in Figure\mref{fig:tradeoff3d}
We have two observations.

First, \system realizes higher attack efficacy than simply combining the adversarial and poisoning attacks. For instance, in the case of \isic (Figure\mref{fig:tradeoff3d}\,(c)), with fidelity loss fixed as 0.2, the adversarial attack achieves $\kappa$ about 0.25; with specificity loss fixed as 0.2, the poisoning attack attains $\kappa$ around 0.4; while \system reaches $\kappa$ above 0.8 under this setting. \ting{This is explained by that \system employs a stronger threat model to jointly optimize the perturbations introduced at both training and inference.}

Second, \system is able to attain attack efficacy unreachable by using each attack vector alone. Across all the cases, \system achieves $\kappa = 1$ under proper fidelity and specificity settings, while the adversarial (or poisoning) attack alone (even with fidelity or specificity loss fixed as 1) is only able to reach $\kappa$ less than 0.9.

\vspace{2pt}
\begin{tcolorbox}[boxrule=0pt, title= Amplification Effect]
  Adversarial inputs and poisoned models amplify each other and give rise to attack efficacy unreachable by using each vector alone.
\end{tcolorbox}

\subsubsection{\bf Empirical Implications}

This amplification effect entails profound implications for the adversary to design more effective attacks. Here we
explore to use adversarial training\mcite{madry:iclr:2018,adv-train-free}, one state-of-the-art defense against adversarial attacks\mcite{Athalye:2018:icml}, to cleanse poisoned models. Starting with the poisoned model, the re-training iteratively updates it with adversarial inputs that deceive its current configuration (i.e., adversarial ``re-training'').


\begin{table}[ht!]{\small
    \centering
    \begin{tabular}{c|c|c}
      \multirow{2}{*}{Dataset}
              & \multicolumn{2}{c}{Maximum Perturbation}                           \\
      \cline{2-3}
              & \pgd                                     & \system                 \\
      \hline
      \cifar  & $3\times \ssup{10}{-2}$                  & $2\times \ssup{10}{-3}$ \\
      \hline
      \imgnet & $4\times \ssup{10}{-3}$                  & $1\times \ssup{10}{-3}$ \\
      \hline
      \isic   & $3\times \ssup{10}{-2}$                  & $1\times \ssup{10}{-3}$ \\
      \hline
      \gtsrb  & $3\times \ssup{10}{-2}$                  & $3\times \ssup{10}{-2}$ \\
    \end{tabular}
    \caption{Maximum input perturbation magnitude for \pgd and \system. \label{tab:magnitude}}}
\end{table}

We perform adversarial re-training on each poisoned model $\atheta$ generated by \system under varying fidelity-specificity trade-off (implementation details in Appendix B). We evaluate the re-trained model $\tbt$ in terms of (i) the attack success rate of \pgd (i.e., $\tbt$'s robustness against regular adversarial attacks), (ii) the attack success rate of $\atheta$'s corresponding adversarial input $\ax$ (i.e., $\tbt$'s robustness against \system), and (iii) $\tbt$'s overall accuracy over benign inputs in the testset. Note that in order to work against the re-trained models, \pgd is enabled with significantly higher perturbation magnitude than \system. Table\mref{tab:magnitude} summarizes \pgd and \system's maximum allowed perturbation magnitude (i.e., fidelity loss) for each dataset.

Observe that adversarial re-training greatly improves the robustness against \pgd, which is consistent with prior work\mcite{madry:iclr:2018,adv-train-free}. Yet, due to the amplification effect, \system retains its high attack effectiveness against the re-trained model. For instance, in the case of \isic (Figure\mref{fig:cross2}\,(c)), even with the maximum perturbation, \pgd attains less than 40\% success rate; in comparison, with two orders of magnitude lower perturbation, \system succeeds with close to 80\% chance. This also implies that adversarial re-training is in general ineffective against \system. Also observe that by slightly increasing the input perturbation magnitude, \system sharply improves the specificity of the poisoned model (e.g., average accuracy over benign inputs), which is attributed to the leverage effect. \ting{Note that while here \system is not adapted to adversarial re-training, it is possible to further optimize the poisoned model by taking account of this defense during training, similar to\mcite{Yao:2019:ccs}.}

\subsection{Analytical Justification}

We now provide analytical justification for the empirical observations regarding the mutual reinforcement effects. 

\subsubsection{\bf Loss Measures}

Without loss of generality, we consider a binary classification setting (i.e., $\mathcal{Y} = \{0, 1\}$), with $(1-t)$ and $t$ being the benign input $\bx$'s ground-truth class and the adversary's target class respectively. Let $f_t(x; \theta)$ be the model $\theta$'s predicted probability that $x$ belongs to $t$. Under this setting, we quantify the set of attack objectives as follows.

\vspace{2pt}
{\em Efficacy --} The attack succeeds only if the adversarial input $\ax$ and poisoned model $\atheta$ force $f_t(\ax; \atheta)$ to exceed 0.5 (i.e., the input crosses the classification boundary).
We thus use $\kappa \triangleq f_t(\bx; \btheta) - 0.5$ to measure the current gap between $\btheta$'s prediction regarding $\bx$ and the adversary's target class $t$.

\vspace{2pt}
{\em Fidelity --}
We quantify the fidelity loss using the $L_p$-norm of the input perturbation:
$\lf(\ax) = \|\ax - \bx\|_p$.
For two adversarial inputs $\ax, \ax'$, we say $\ax < \ax'$ if
$\lf(\ax) < \lf(\ax')$.
For simplicity, we use $p = 2$, while the analysis generalizes to other norms as well.

As shown in Figure\mref{fig:leverage}\,(a), in a successful adversarial attack (with the adversarial input $\rbx$), if the perturbation magnitude is small enough, we can approximate the fidelity loss as $\bx$'s distance to the classification boundary\mcite{Dezfooli:2017:arxiv}:
$\lf(\rbx) \approx \kappa/\|\ssub{\nabla}{\rx} \ell(\bx;\btheta)\|_2$,
where a linear approximation is applied to the loss function. In the following, we denote $h \triangleq \lf(\rbx)$.

\vspace{2pt}
{\em Specificity --}
Recall that the poisoned model $\atheta$ modifies $\bx$'s surrounding classification boundary, as shown in Figure\mref{fig:leverage} (b). While it is difficult to exactly describe the classification boundaries encoded by {\dnns}\mcite{Fawzi:2017:arxiv}, we approximate the local boundary surrounding an input with the surface of a $d$-dimensional sphere, where $d$ is the input dimensionality. This approximation is justified as follows.

First, it uses a quadratic form, which is more expressive than a linear approximation\mcite{Dezfooli:2017:arxiv}. Second, it reflects the impact of model complexity on the boundary: the maximum possible curvature of the boundary is often determined by the model's inherent complexity\mcite{Fawzi:2017:arxiv}. For instance, the curvature of a linear model is 0, while a one hidden-layer neural network with an infinite number of neurons is able to model arbitrary boundaries\mcite{Cybenko:mcss:1989}. We relate the model's complexity to the maximum possible curvature, which corresponds to the minimum possible radius of the sphere.

The boundaries before and after the attacks are thus described by two hyper-spherical caps. As the boundary before the attack is fixed, without loss of generality, we assume it to be flat for simplicity. Now according to \meq{eq:sloss}, the specificity loss is measured by the number of inputs whose classifications are changed due to $\rt$. Following the assumptions, such inputs reside in a $d$-dimensional hyper-spherical cap, as shown in Figure\mref{fig:leverage}\,(b). 
Due to its minuscule scale, the probability density $p_\mathrm{data}$ in this cap is roughly constant. Minimizing the specificity loss is thus equivalent to minimizing the cap volume\mcite{Polyanin:2006}, which amounts to maximizing the curvature of the sphere (or minimizing its radius). Let $r$ be the minimum radius induced by the model. We quantify the specificity loss as:
\begin{equation}
  \label{eq:sloss2}
  \ls(\rt) = p_\mathrm{data}  \frac{\pi^{\frac{d-1}{2}}r^d }{\Gamma \left( \frac{d+1}{2}\right)} \int_{0}^{\arccos \left(1-\frac{h}{r} \right)}\sin^d (t)\,\mathrm{d}t
\end{equation}
where $\Gamma(z) \triangleq \int_0^\infty t^{z-1}e^{-t}\,\mathrm{d}t$ is the Gamma function.

\subsubsection{\bf Mutual Reinforcement Effects}

Let $\rbx,\rbt$ be the adversarial input and poisoned model given by the adversarial and poisoning attacks respectively, and $(\ax, \atheta)$ be the adversarial input and poisoned model generated by \system. Note that for fixed attack efficacy, $\rxs = \rbx$ if $\rts = \btheta$ and $\rts = \rbt$ if $\rxs = \bx$.

\vspace{3pt}
{\em Leverage Effect --}
We now quantify the leverage effect in the case of trading fidelity for specificity, while the alternative case can be derived similarly. Specifically, this effect is measured by the ratio of specificity ``saving'' versus fidelity ``cost'', which we term as the {\em leverage effect coefficient}:
\begin{equation}
  \label{eq:le}
  \phi(\rxs, \rts) \triangleq \frac{ 1 -\ls (\rts)/\ls (\rbt) }{ \lf (\rxs)  / \lf (\rbx) }
\end{equation}

Intuitively, the numerator is the specificity ``saving'', while the denominator is the fidelity ``cost''. We say that the trade-off is significantly disproportionate, if $\phi(\rxs, \rts) \gg 1$, i.e., the saving dwarfs the cost. It is trivial to verify that if $\phi(\rxs, \rts) \gg 1$ then the effect of trading specificity for fidelity is also significant
$\phi(\rts, \rxs) \gg 1$.\footnote{If $(1-x)/y \gg 1$ then $(1-y)/x \gg 1$ for $0 < x, y < 1$.}

Consider the \system attack as shown in Figure\mref{fig:leverage} (c). The adversarial input $\ax$ moves towards the classification boundary and reduces the loss by $\kappa' (\kappa' < \kappa)$. The perturbation magnitude is thus at least $\kappa'/\|\ssub{\nabla}{\rx} \ell(\bx;\btheta)\|_2$. The relative fidelity loss is given by:
\begin{equation}
  \label{eq:le0}
  \lf (\rxs)  /\lf (\rbx) = \kappa'/\kappa
\end{equation}
Below we use $z = \kappa'/\kappa$ for a short notation.

Meanwhile, it is straightforward to derive that the height of the hyper-spherical cap is $(1 - z) h$. The relative specificity loss is thus:
\begin{equation}
  \label{eq:le2}
  \ls (\rts)/ \ls (\rbt) = \frac{\int_{0}^{\arccos \left(1 - \frac{h}{r} + z\frac{h}{r} \right)}\sin^d (t)\,\mathrm{d}t}{\int_{0}^{\arccos \left(1 - \frac{h}{r} \right)}\sin^d (t)\,\mathrm{d}t}
\end{equation}

Instantiating \meq{eq:le} with \meq{eq:le0} and \meq{eq:le2}, the leverage effect of trading fidelity for specificity is defined as:
\begin{empheq}{equation}
  \label{eq:le3}
  \phi(\rxs, \rts) = \frac{\int^{\arccos \left(1 - \frac{h}{r} \right)}_{\arccos \left(1 - \frac{h}{r} + z\frac{h}{r} \right)}\sin^d (t)\,\mathrm{d}t}{z \int_{0}^{\arccos \left(1 - \frac{h}{r} \right)}\sin^d (t)\,\mathrm{d}t}
\end{empheq}

The following proposition justifies the effect of trading fidelity for specificity (proof in Appendix A). A similar argument can be derived for trading specificity for fidelity.
\begin{proposition}
  \label{prop:leverage}
  The leverage effect defined in \meq{eq:le3} is strictly greater than 1 for any $0 < z < 1$.
\end{proposition}

Intuitively, to achieve fixed attack efficacy ($\kappa$), with a slight increase of fidelity loss $\lf(\rxs)$, the specificity loss $\ls(\rts)$ is reduced super-linearly.

\begin{figure}[!ht]
  \centering
  \epsfig{file = 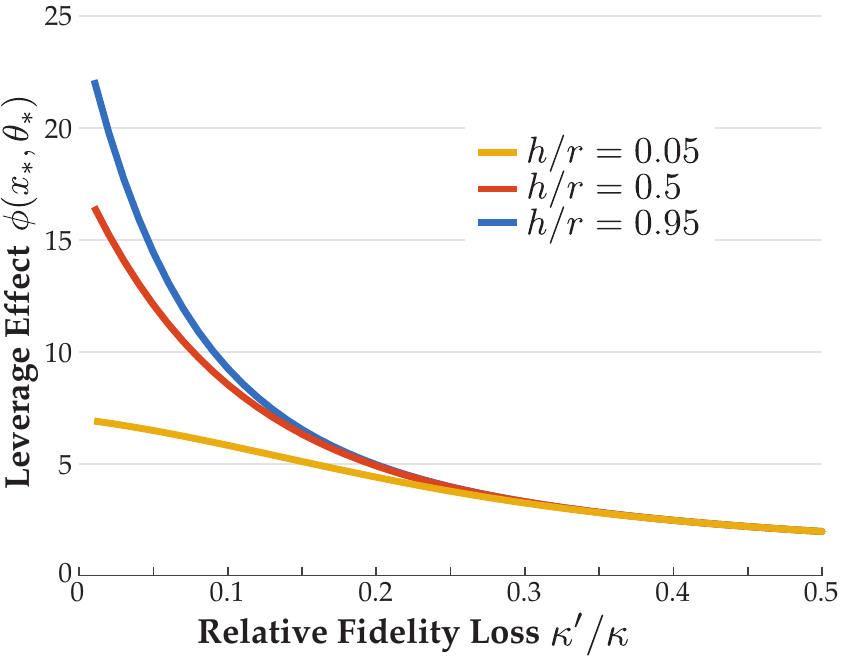, width = 55mm}
  \caption{Leverage effect with respect to the relative fidelity loss $z$ and the minimum radius $r$ (with $d = 50$). \label{fig:leverage3}}
\end{figure}

Figure\mref{fig:leverage3} evaluates this effect as a function of relative fidelity loss under varying setting of $h/r$. Observe that the effect is larger than 1 by a large margin, especially for small fidelity loss $\kappa'/\kappa$, which is consistent with our empirical observation: with little fidelity cost, it is possible to significantly reduce the specificity loss.


\vspace{3pt}
{\em Amplification Effect --} From Proposition\mref{prop:leverage}, we can also derive the explanation for the amplification effect.

Consider an adversarial input $\ax$ that currently achieves attack efficacy $\kappa'$ with relative fidelity loss $\kappa'/\kappa$. Applying the poisoned model $\atheta$ with relative specificity loss $(1 -\kappa'/\kappa)/\phi(\ax, \atheta)$, the adversary is able to attain attack efficacy $\kappa$. In other words, the poisoned model $\atheta$ ``amplifies'' the attack efficacy of the adversarial input $\ax$ by $\kappa/\kappa'$ times, with cost much lower than required by using the adversarial attack alone to reach the same attack efficacy (i.e., $1 - \kappa'/\kappa$), given that $\phi(\ax, \atheta) \gg 1$ in Proposition\mref{prop:leverage}.

%% file: study3.tex
\section{IMC-Optimized Attacks}
\label{sec:study3}

In this section, we demonstrate that \system, as a general attack framework, can be exploited to enhance existing attacks with respect to multiple metrics. We further discuss potential countermeasures against such optimized attacks and their technical challenges.

\subsection{Attack Optimization}

\subsubsection{\bf Basic Attack}
We consider {\tnet}\mcite{Liu:2018:ndss}, a representative backdoor attack, as the reference attack model. At a high level, \tnet defines a specific pattern (e.g., watermark) as the trigger and enforces the poisoned model to misclassify all the inputs embedded with this trigger. \ting{As it optimizes both the trigger and poisoned model, \tnet enhances other backdoor attacks (e.g., {\bnet}\mcite{Gu:arxiv:2017}) that employ fixed trigger patterns.} 

Specifically, the attack consists of three steps. (i) First, the trigger pattern is partially defined in an ad hoc manner; that is, the watermark shape (e.g., square) and embedding position are pre-specified. (ii) Then, the concrete pixel values of the trigger are optimized to activate neurons rarely activated by benign inputs, in order to minimize the impact on benign inputs. (iii) Finally, the model is re-trained to enhance the effectiveness of the trigger pattern. 

Note that within \tnet, the operations of trigger optimization and model re-training are executed independently. It is thus possible that after re-training, the neurons activated by the trigger pattern may deviate from the originally selected neurons, resulting in suboptimal trigger patterns and/or poisoned models. \ting{Also note that \tnet works without access to the training data; yet, to make fair comparison, in the following evaluation, \tnet also uses the original training data to construct the backdoors.}

\subsubsection{\bf Enhanced Attacks} We optimize \tnet within the \system framework. \ting{Compared with optimizing the trigger only\mcite{cnn-trigger-generation}, \system improves \tnet in terms of both attack effectiveness and evasiveness.} Specifically, let $r$ denote the trigger. We initialize $r$ with the trigger pre-defined by \tnet and optimize it using the co-optimization procedure. To this end, we introduce a mask $m$ for the given benign input $\bx$. For $\bx$'s $i$-th dimension (pixel), we define $m[i] = 1- p$ ($p$ is the transparency setting) if $i$ is covered by the watermark and $m[i] = 0$ otherwise. Thus the perturbation operation is defined as $\ax = \psi(\bx, r;m) = \bx \odot (1-m) + r \odot m$, where $\odot$ denotes element-wise multiplication. We reformulate the backdoor attack in \meq{eq:opt5} as follows:
\begin{equation}
\label{eq:opt7}
\min_{\rt, r, m} \,\, \sE_{\bx \in \calT}\, [\ell(\psi(\bx, r;m), \ay; \theta)] + \lambda \lf (m) + \nu \ls (\rt)
\end{equation}
where we define the fidelity loss in terms of $m$. Typically, $\lf(m)$ is defined as $m$'s $L_1$ norm and $\ls (\rt)$ is the accuracy drop on benign cases similar to \meq{eq:sloss}. 

Algorithm\mref{alg:watermark} sketches the optimization procedure of \meq{eq:opt7}. It alternates between optimizing the trigger and mask (line 4) and optimizing the poisoned model (line 5). Specifically, during the trigger perturbation step, we apply the Adam optimizer\mcite{kingma:iclr:2015}. Further, instead of directly optimizing $r$ which is bounded by $[0, 1]$, we apply change-of-variable and optimize over a new variable $w_r \in (-\infty,+\infty)$, such that $r = (\tanh(w_r)+1)/2$ (the same trick is also applied on $m$). Note that Algorithm\mref{alg:watermark} represents a general optimization framework, which is adaptable to various settings. For instance, one may specify all the non-zero elements of $m$ to share the same transparency or optimize the transparency of each element independently (details in \msec{sec:human} and \msec{sec:tool}). 
In the following, we term the enhanced \tnet as \ntnet. 

\begin{algorithm}[!ht]{\small
\KwIn{initial trigger mask -- $\ssub{m}{\circ}$; benign model -- $\btheta$; target class -- $\ay$; hyper-parameters -- $\lambda, \nu$}
\KwOut{trigger mask -- $m$; trigger pattern -- $r$, poisoned model -- $\atheta$}
\tcp{\footnotesize initialization}
$\ssup{\theta}{(0)}, k \leftarrow \btheta, 0$\;
$\ssup{m}{(0)}, \ssup{r}{(0)} \leftarrow m_0, \tnet(m_0)$\;
\tcp{\footnotesize optimization}
\While{not converged yet}{
\tcp{\footnotesize trigger perturbation}
$\ssup{r}{(k+1)}, \ssup{m}{(k+1)} \leftarrow \arg\min_{r, m} \sE_{\bx \in \calT}\, [\ell(\psi(\bx, r;m), \ay; \ssup{\theta}{(k)})] + \lambda \lf (m)$\;
\tcp{\footnotesize model perturbation}
$\ssup{\theta}{(k+1)} \leftarrow \arg\min_{\theta} \sE_{\bx \in \calT}\,[\ell(\psi(\bx, \ssup{r}{(k)};\ssup{m}{(k)}), \ay; \theta)] + \nu \ls(\theta)$\;
$k \leftarrow k + 1$\;
}
\Return $(\ssup{m}{(k)}, \ssup{r}{(k)}, \ssup{\theta}{(k)})$\;
\caption{\ntnet Attack \label{alg:watermark}}}
\end{algorithm}

\subsection{Optimization against Human Vision}
\label{sec:human}

We first show that \ntnet is optimizable in terms of its evasiveness with respect to human vision. The evasiveness is quantified by the size and transparency (or opacity) of trigger patterns. Without loss of generality, we use square-shaped triggers. The trigger size is measured by the ratio of its width over the image width.

\begin{figure}[ht!]
  \centering
  \epsfig{file = 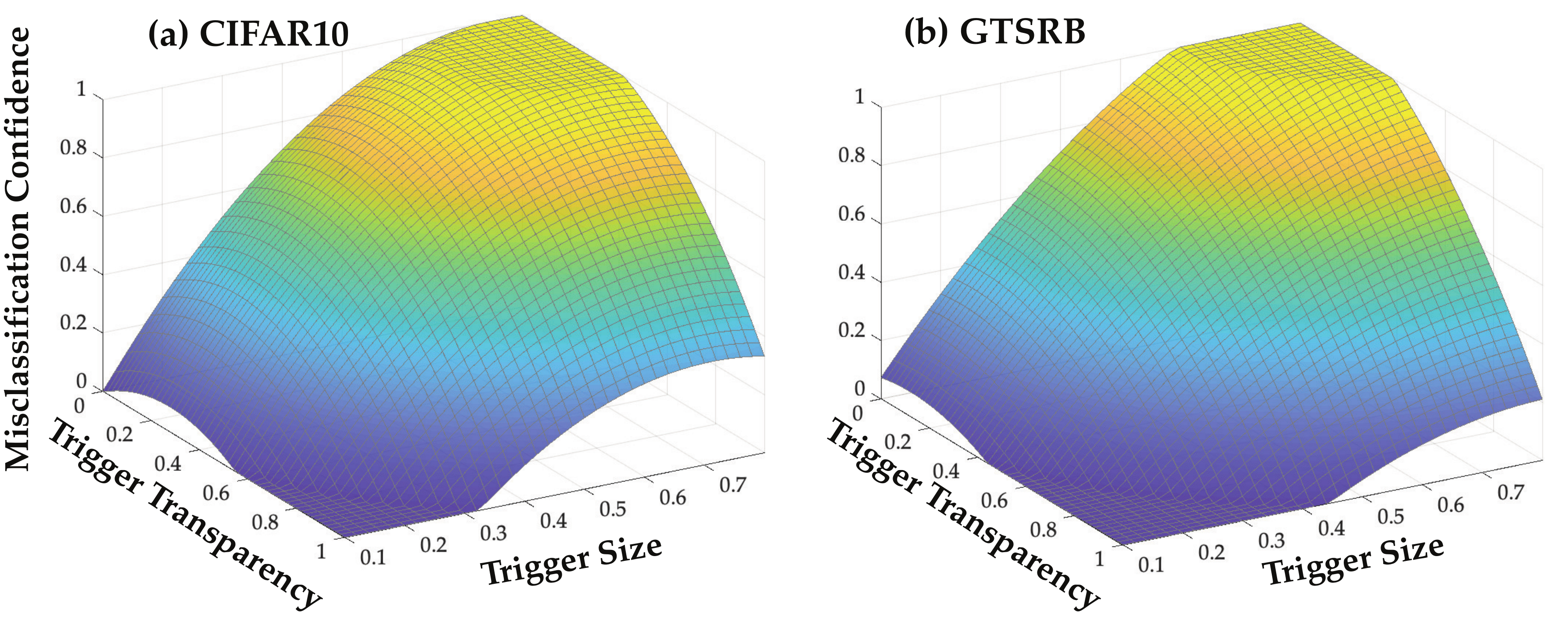, width = 80mm}
  \caption{Attack efficacy of \ntnet as a function of trigger size and transparency. \label{fig:watermark3d}}
\end{figure}

Figure\mref{fig:watermark3d} illustrates \ntnet's attack efficacy (average misclassification confidence of trigger-embedded inputs) under varying evasiveness constraints. Observe that the efficacy increases sharply as a function of the trigger size or opacity. Interestingly, the trigger size and opacity also demonstrate strong mutual reinforcement effects: (i) leverage - for fixed attack efficacy, by a slight increase in opacity (or size), it significantly reduces the size (or opacity); (ii) amplification - for fixed opacity (or size), by slightly increasing size (or opacity), it greatly boosts the attack efficacy.

\begin{figure}[ht!]
  \centering
  \epsfig{file = 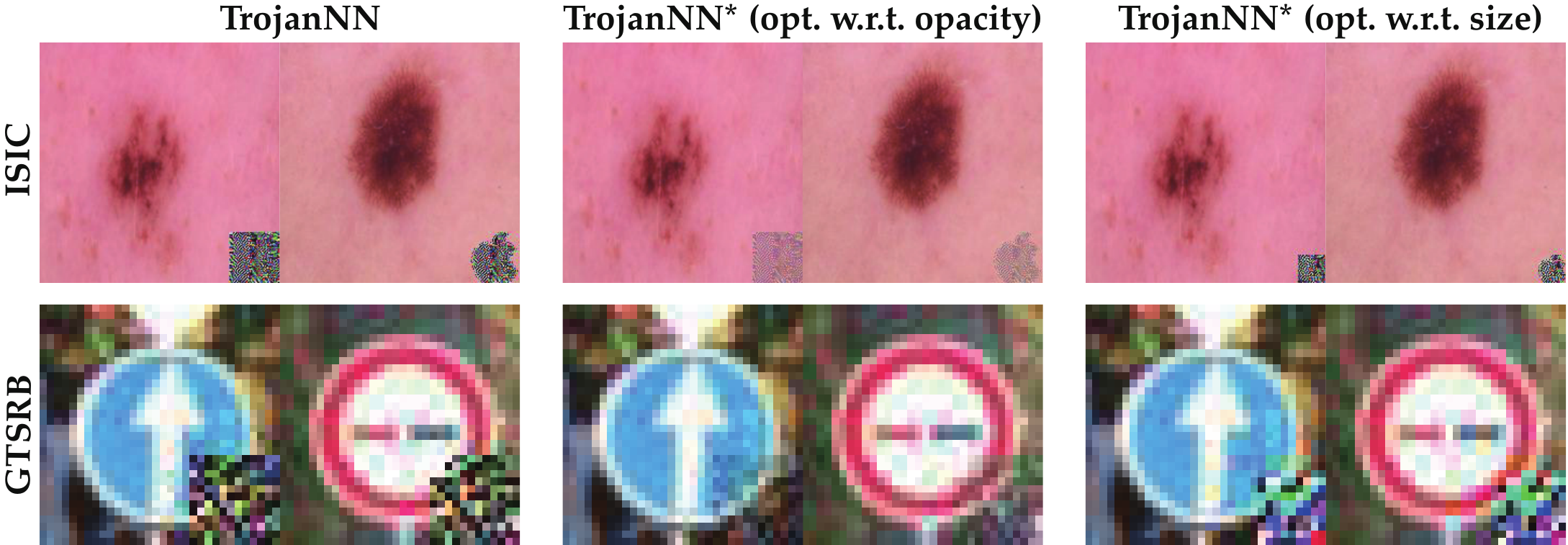, width = 80mm}
  \caption{Sample triggers generated by \tnet (a), \ntnet optimizing opacity (b) and optimizing size (c). \label{fig:trigger}}
\end{figure}

To further validate the leverage effect, we compare the triggers generated by \tnet and \ntnet. Figure\mref{fig:trigger} shows sample triggers given by \tnet and \ntnet under fixed attack efficacy (with $\kappa = 0.95$). It is observed that compared with \tnet, \ntnet significantly increases the trigger transparency (under fixed size) or minimizes the trigger size (under fixed opacity).

\begin{figure*}[ht!]
  \centering
  \epsfig{file = 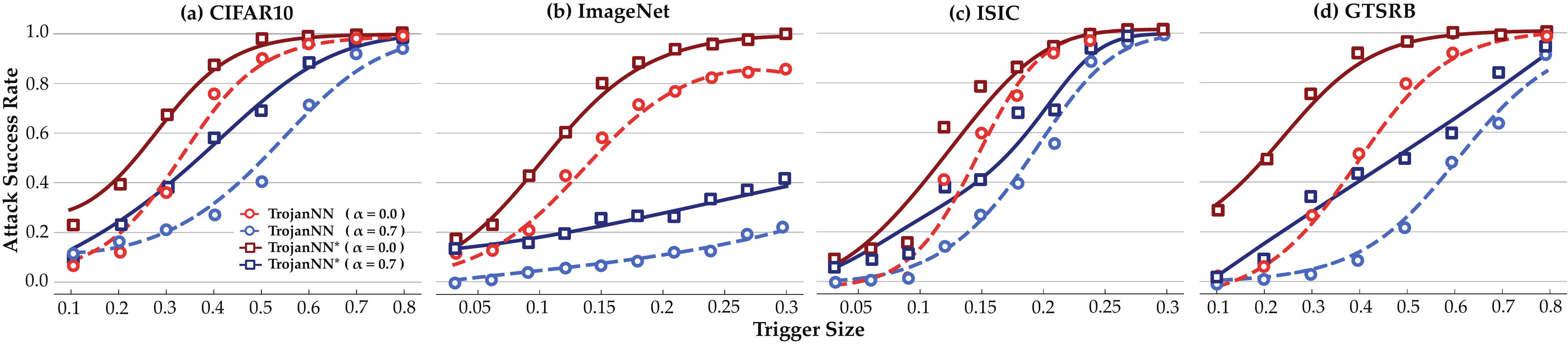, width = 170mm}
  \caption{ASR of \tnet and \ntnet as functions of trigger size. \label{fig:watermark_size}}
  \end{figure*}

\begin{figure*}[ht!]
  \centering
  \epsfig{file = 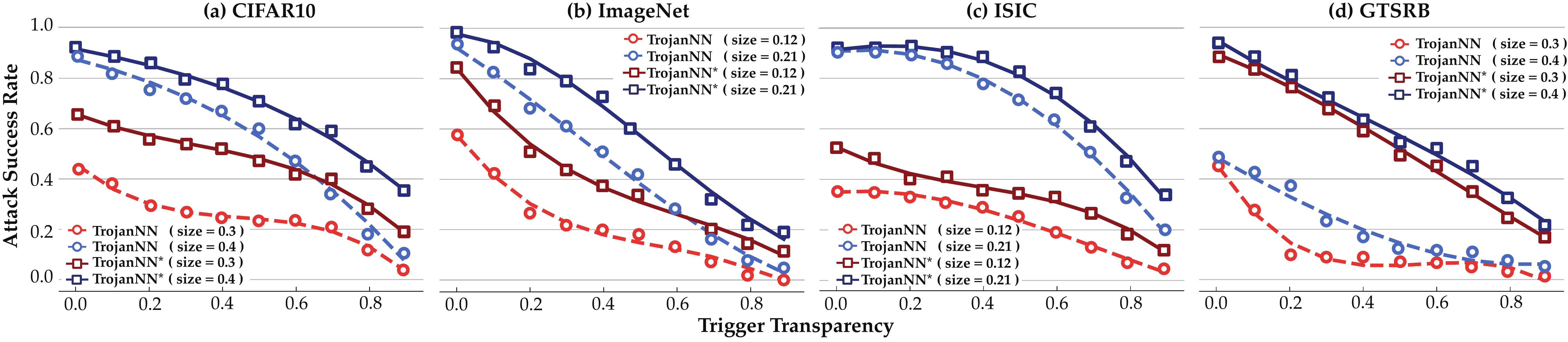, width = 170mm}
  \caption{ASR of \tnet and \ntnet as functions of trigger transparency. \label{fig:watermark_alpha}}
  \end{figure*}

To further validate the amplification effect, we measure the attack success rate (ASR) of \tnet and \ntnet under varying evasiveness constraints (with $\kappa = 0.95$), with results shown in Figure\mref{fig:watermark_size} and\mref{fig:watermark_alpha}. It is noticed that across all the datasets, \ntnet outperforms \tnet by a large margin under given trigger size and opacity. For instance, in the case of \gtsrb (Figure\mref{fig:watermark_size}\,(d)), with trigger size and transparency fixed as 0.4 and 0.7, \ntnet outperforms \tnet by 0.39 in terms of ASR; in the case of \cifar (Figure\mref{fig:watermark_alpha}\,(a)), with trigger size and transparency fixed as 0.3 and 0.2, the ASRs of \ntnet and \tnet differ by 0.36.

We can thus conclude that leveraging the co-optimization framework, \ntnet is optimizable with respect to human detection without affecting its attack effectiveness.

\subsection{Optimization against Detection Methods}
\label{sec:tool}

In this set of experiments, we demonstrate that \ntnet is also optimizable in terms of its evasiveness with respect to multiple automated detection methods.

\subsubsection{\bf Backdoor Detection}
The existing backdoor detection methods can be roughly classified in two categories based on their application stages and detection targets. The first class is applied at the model inspection stage and aims to detect suspicious models and potential backdoors\mcite{Wang:2019:sp,Chen:2019:IJCAI,abs}; the other class is applied at inference time and aims to detect trigger-embedded inputs\mcite{Chen:2018:arxiv,Chou:2018:arxiv,Gao:2019:arxiv,Doan:2020:arxiv}. In our evaluation, we use {\nc}\mcite{Wang:2019:sp} and {\strip}\mcite{Gao:2019:arxiv} as the representative methods of the two categories. In Appendix C, we also evaluate \tnet and \ntnet against {\abs}\mcite{abs}, another state-of-the-art backdoor detector.

\vspace{2pt}
{\em NeuralCleanse --} For a given \dnn, \nc searches for potential triggers in every class. Intuitively, if a class is embedded with a backdoor, the minimum perturbation (measured by its $L_1$-norm) necessary to change all the inputs in this class to the target class is abnormally smaller than other classes. \another{Empirically, after running the trigger search algorithm over 1,600 randomly sampled inputs for 10 epochs, a class with its minimum perturbation normalized by median absolute deviation exceeding 2.0 is considered to contain a potential backdoor with 95\% confidence.}

\vspace{2pt}
{\em STRIP --} 
For a given input, \strip mixes it up with a benign input \another{using equal weights}, feeds the mixture to the target model, and computes the entropy of the prediction vector (i.e., self-entropy). Intuitively, if the input is embedded with a trigger, the mixture is still dominated by the trigger and tends to be misclassified to the target class, resulting in relatively low self-entropy; otherwise, the self-entropy tends to be higher. \another{To reduce variance, for a given input, we average its self-entropy with respect to 8 randomly sampled benign inputs. We set the positive threshold as 0.05 and measure \strip's effectiveness using F-1 score.}

\begin{figure*}[ht!]
  \centering
  \epsfig{file = 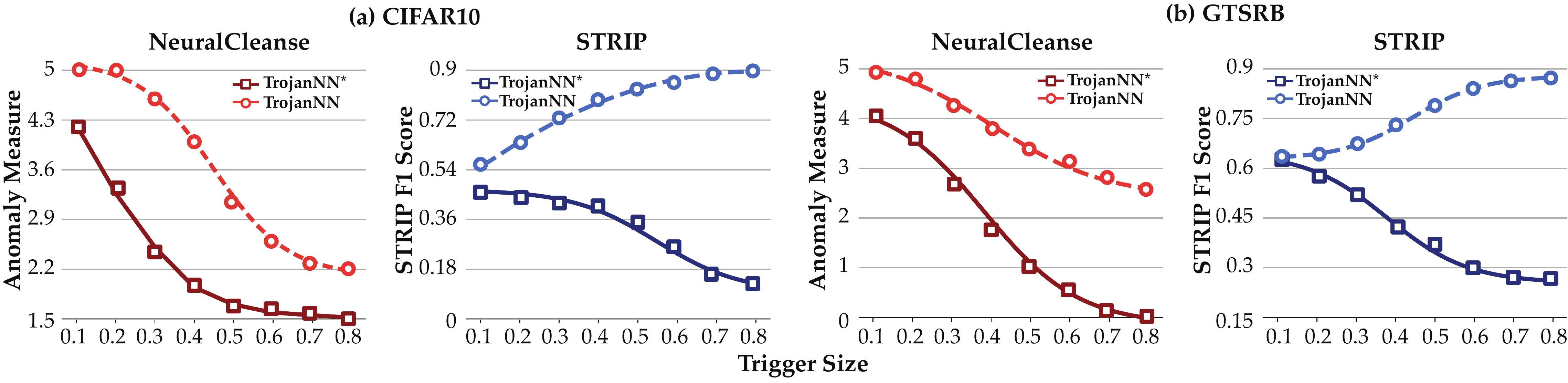, width = 170mm}
  \caption{Detection of \tnet and \ntnet by \nc and \strip on \cifar and \gtsrb. \label{fig:watermark_detect}}
\end{figure*}

\subsubsection{\bf Attack Optimization} We optimize \ntnet in terms of its evasiveness with respect to both \nc and \strip. Both detectors aim to detect anomaly under certain metrics, which we integrate into the loss terms in Algorithm\mref{alg:watermark}. 

Specifically, \nc searches for potential trigger with minimum $L_1$-norm, which is related to the mask $m$. We thus instantiate the fidelity loss $\lf(m)$ as $m$'s $L_1$-norm and optimize it during the trigger perturbation step. To normalize  $\lf(m)$ to an appropriate scale, we set the hyper-parameter $\lambda$ as the number of pixels covered by the trigger. Meanwhile, \strip mixes adversarial and benign inputs and computes the self-entropy of the mixtures, which highly depends on the model's behaviors. We thus instantiate the specificity loss $\ls(\theta)$ as $\sE_{x, x' \in \gR}\, [-H (f(\frac{\ssub{x}{*}}{2}+\frac{x'}{2};\theta))]$, in which we randomly mix up an adversarial input $\ssub{x}{*}$ (via perturbing a benign input $x$) and another benign input $x'$ and maximize the self-entropy of their mixture.


\subsubsection{\bf Detection Evasiveness}
We apply the above two detectors to detect \tnet and \ntnet, with results summarized in Figure\mref{fig:watermark_detect}. We have the following observations. First, the two detectors are fairly effective against \tnet. In comparison, \ntnet demonstrates much higher evasiveness. For instance, in the case of \gtsrb (Figure\mref{fig:watermark_detect}\,(b)), with trigger size fixed as 0.4, the anomaly measures of \ntnet and \tnet by \nc differ by over 2, while the F-1 scores on \ntnet and \tnet by \strip differ by more than 0.3. We thus conclude that \ntnet is optimizable in terms of evasiveness with respect to multiple detection methods simultaneously.

\subsection{Potential Countermeasures}

Now we discuss potential mitigation against \system-optimized attacks and their technical challenges. It is shown above that using detectors against adversarial inputs or poisoned models independently is often insufficient to defend against \system-optimized attacks, due to the mutual reinforcement effects. One possible solution is to build ensemble detectors that integrate individual ones and detect \system-optimized attacks based on both input and model anomaly.


To assess the feasibility of this idea, we build an ensemble detector against \ntnet via integrating \nc and \strip. Specifically, we perform the following detection procedure: (i) applying \nc to identify the potential trigger, (ii) for a given input, attaching the potential trigger to a benign input, (iii) mixing this benign input up with the given input under varying mixture weights, (iv) measuring the self-entropy of these mixtures, and \ting{(v)} using the standard deviation of the self-entropy values to distinguish benign and trigger-embedded inputs.

Intuitively, if the given input is trigger-embedded, the mixture combines two trigger-embedded inputs and is thus dominated by one of the two triggers, regardless of the mixture weight, resulting in a low deviation of self-entropy. In comparison, if the given input is benign, the mixture is dominated by the trigger only if the weight is one-sided, resulting in a high deviation of self-entropy.

\begin{figure}[ht!]
  \centering
  \epsfig{file = 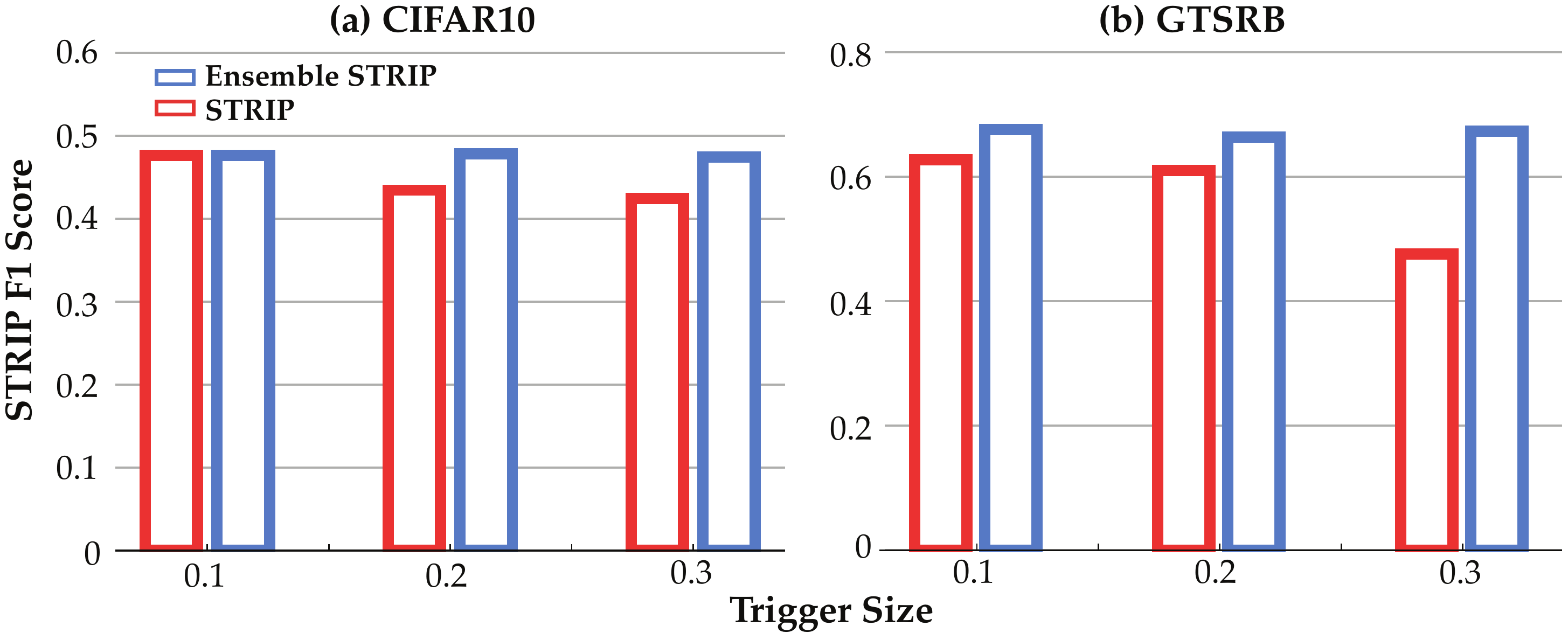, width = 80mm}
  \caption{Detection of basic and ensemble \strip against \ntnet on \cifar and \gtsrb.\label{fig:detect_new_bar}}
  \end{figure}

We compare the performance of the basic and ensemble \strip against \ntnet \ting{(the detection against \tnet is deferred to Appendix C)}. As shown in Figure\mref{fig:detect_new_bar}, the ensemble detector performs slightly better across all the cases, implying the effectiveness of the ensemble approach. However, the improvement is marginal (less than 0.2), especially in the case of small-sized triggers. This may be explained by the inherent challenges of defending against \system-optimized attacks: due to the mutual reinforcement effects, \ntnet attains high attack efficacy with minimal input and model distortion; it thus requires to carefully account for such effects in order to design effective countermeasures.

%% file: literature.tex
\section{Related Work}
\label{sec:literature}

With their increasing use in security-sensitive domains, DNNs are becoming the new targets of malicious manipulations\mcite{Biggio:2018:pr}. Two primary attack vectors have been considered in the literature: adversarial inputs and poisoned models.


\subsubsection*{\bf Adversarial Inputs} The existing research on adversarial inputs is divided in two campaigns. 

One line of work focuses on developing new attacks against {\dnns}\mcite{szegedy:iclr:2014,goodfellow:fsgm,papernot:eurosp:2017,carlini:sp:2017}, with the aim of crafting adversarial samples to force \dnns to misbehave. The existing attacks can be categorized as untargeted (in which the adversary desires to simply force misclassification) and targeted (in which the adversary attempts to force the inputs to be misclassified into specific classes).

Another line of work attempts to improve \dnn resilience against adversarial attacks by devising new training strategies (e.g., adversarial training)\mcite{papernot:sp:2016,kurakin:advbim,guo:iclr:2018,Tramer:2018:iclr} or detection mechanisms\mcite{Meng:2017:ccs,Xu:2018:ndss,Gehr:2018:sp,Ma:2019:ndss}. However, the existing defenses are often penetrated or circumvented by even stronger attacks\mcite{Athalye:2018:icml,Ling:2019:sp}, resulting in a constant arms race between the attackers and defenders.

\subsubsection*{\bf Poisoned Models} The poisoned model-based attacks can be categorized according to their target inputs. In the poisoning attacks, the target inputs are defined as non-modified inputs, while the adversary's goal is to force such inputs to be misclassified by the poisoned {\dnns}\mcite{Ji:2017:cns,Ji:2018:ccsa,Shafahi:2018:nips,Wang:2018:sec,Suciu:2018:sec}. In the backdoor attacks, specific trigger patterns (e.g., a particular watermark) are pre-defined, while the adversary's goal is to force any inputs embedded with such triggers to be misclassified by the poisoned models\mcite{Liu:2018:ndss,Gu:arxiv:2017}. Note that compared with the poisoning attacks, the backdoor attacks leverage both adversarial inputs and poisoned models.


The existing defense methods against poisoned models mostly focus on the backdoor attacks, which, according to their strategies, can be categorized as: (i) cleansing potential contaminated data at the training stage\mcite{Tran:2018:nips}, (ii) identifying suspicious models during model inspection\ting{\mcite{Wang:2019:sp,Chen:2019:IJCAI,abs}}, and (iii) detecting trigger-embedded inputs at inference time\mcite{Chen:2018:arxiv,Chou:2018:arxiv,Gao:2019:arxiv,Doan:2020:arxiv}.


\vspace{3pt}
Despite the intensive research on adversarial inputs and poisoned models in parallel, there is still a lack of understanding about their inherent connections. This work bridges this gap by studying the two attack vectors within a unified framework and providing a holistic view of the vulnerabilities of \dnns deployed in practice.

%% file: conclusion.tex
\section{Conclusion}
\label{sec:conclusion}

This work represents a solid step towards understanding adversarial inputs and poisoned models in a unified manner. We show both empirically and analytically that (i) there exist intriguing mutual reinforcement effects between the two attack vectors, (ii) the adversary is able to exploit such effects to optimize attacks with respect to multiple metrics, and (iii) it requires to carefully account for such effects in designing effective countermeasures against the optimized attacks. We believe our findings shed light on the holistic vulnerabilities of \dnns deployed in realistic settings.

This work also opens a few avenues for further investigation. First, besides the targeted, white-box attacks considered in this paper, it is interesting to study the connections between the two vectors under alternative settings (e.g., untargeted, black-box attacks).
Second, enhancing other types of threats (e.g., latent backdoor attacks) within the input-model co-optimization framework is  a direction worthy of exploration. Finally, devising a unified robustness metric accounting for both vectors may serve as a promising starting point for developing effective countermeasures.

%% file: appendix.tex
\section*{Appendix}

\subsection*{\bf A. Proofs}

\subsubsection*{\bf A0. Preliminaries}

In the following proofs, we use the following definitions for notational simplicity:
\begin{empheq}[box=\fbox]{align}
\nonumber  \alpha \triangleq h/r
\end{empheq}
\begin{empheq}[box=\fbox]{align}
\nonumber  y \triangleq 1 - z
\end{empheq}
Further we have the following result.
\begin{empheq}[box=\fbox]{align}
  \label{eq:fact}
  \int_{0}^{\arccos{\left(x\right)}}\sin^d{\left(t\right)}\mathrm{d}t
 = \int_{x}^{1}\left(1-t^2\right)^\frac{d-1}{2}\mathrm{d}t
\end{empheq}

\subsubsection*{\bf A1. Proof of Proposition\mref{the:danskin}}

\begin{proof}
Recall that $\ssub{\gF}{\epsilon}(\bx)$ represents a non-empty compact set, $\ell(\rx; \cdot)$ is differentiable for $\rx \in \ssub{\gF}{\epsilon}(\bx)$, and $\nabla_{\rt} \ell(\rx, \rt)$ is continuous over the domains $\ssub{\gF}{\epsilon}(\bx)  \times \sR^n$.

\vspace{2pt}
Let $ \sboth{\gF}{\epsilon}{*}(\bx) =  \{\arg\min_{\rx \in \ssub{\gF}{\epsilon}(\bx)} \ell(\rx; \rt)\}$ be the set of minimizers and $\ell(\rt) \triangleq \min_{\rx \in \ssub{\gF}{\epsilon}(\bx)} \ell(\rx; \rt)$. The Danskin's theorem~\cite{Danskin:book} states that $\ell(\rt)$ is locally continuous and directionally differentiable. The derivative of $\ell(\rt)$ along the direction $d$ is given by 
\begin{align}
  \nonumber
\mathrm{D}_d \ell(\rt) = \min_{\rx \in \sboth{\gF}{\epsilon}{*}(\bx)} \ssup{d}{\top} \nabla_{\rt} \ell(\rx, \rt)
\end{align}

\vspace{2pt}
We apply the Danskin's theorem to our case. Let $\rx^* \in  \mathcal{F}_\epsilon(\bx)$ be a minimizer of $\min_{\rx} \ell(\rx; \rt)$. Consider the direction $d = -\nabla_{\rt} \ell(\rx^*; \rt)$. We then have:
\begin{align}
\nonumber
\mathrm{D}_d \ell(\rt) & = \min_{\rx \in \sboth{\gF}{\epsilon}{*}(\bx)} \ssup{d}{\top} \nabla_{\rt} \ell(\rx, \rt) \\
\nonumber
& \leq -\|  \nabla_{\rt} \ell(\rx^*, \rt) \|_2^2 \leq 0
\end{align}

Thus, it follows that $\nabla_{\rt} \ell(\rx^*; \rt)$ is a proper descent direction of $\min_{\rx \in \mathcal{F}_\epsilon(\bx)} \ell (\rx; \rt)$.
\end{proof}

Note that in the proof above, we ignore the constraint of $\rt \in \ssub{\gF}{\delta}(\btheta)$. Nevertheless, the conclusion is still valid. With this constraint, instead of considering the global optimum of $\rt$, we essentially consider its local optimum within $\ssub{\gF}{\delta}(\btheta)$. Further, for DNNs that use constructs such as ReLU, the loss function is not necessarily continuously differentiable. However, since the set of discontinuities has measure zero, this is not an issue in practice.

\subsubsection*{\bf A2. Proof of Proposition\mref{prop:leverage}}

\begin{proof} Proving $\phi(\rxs, \rts) > 1$ is equivalent to showing the following inequality:
\begin{displaymath}
\frac{\int_{0}^{\arccos{\left(1-y\alpha\right)}}\sin^d{\left(t\right)}\mathrm{d}t}{y}<\int_{0}^{\arccos{\left(1-\alpha\right)}}\sin^d{\left(t\right)}\mathrm{d}t
\end{displaymath}

We define $f(y)=\frac{1}{y} \int_{0}^{\arccos \left(1-y\alpha\right)} \sin^d (t) \mathrm{d}t$. This inequality is equivalent to
  $f(y)<f(1)$ for $y \in (0, 1)$. It thus suffices to prove $f'(y) > 0$.

  Considering \meq{eq:fact}, we have $f(y) = \frac{1}{y}\int_{1-y\alpha}^1 \left(1-t^2\right)^\frac{d-1}{2}\mathrm{d}t$
and $f'(y) =  g(y)/y^2$ where
\begin{displaymath}
g(y) =  y\alpha\left(1-\left(1-y\alpha\right)^2\right)^\frac{d-1}{2}-\int_{1-y\alpha}^{1}\left(1-t^2\right)^\frac{d-1}{2}\mathrm{d}t
\end{displaymath}

Denote $x = 1 - y\alpha$. We have
\begin{displaymath}
g(x) = \left(1+x\right)^\frac{d-1}{2}\left(1-x\right)^\frac{d+1}{2}-\int_{x}^{1}\left(1-t^2\right)^\frac{d-1}{2}\mathrm{d}t
\end{displaymath}

Note that $g(1) = 0$. With $d > 1$, we have
\begin{displaymath}
g'(x) = -\left(d-1\right)x\left(1+x\right)^\frac{d-3}{2}\left(1-x\right)^\frac{d-1}{2}<0
\end{displaymath}

Therefore, $g(x) > 0$ for $x \in (0, 1)$, which in turn implies $f'(y) > 0$ for $y \in (0, 1)$.
\end{proof}

\subsection*{\bf B. Implementation Details}

Here we elaborate on the implementation of attacks and defenses in this paper. 

\subsubsection*{\bf B1. Parameter Setting}

Table\mref{tab:param} summarizes the default parameter setting in our empirical evaluation (\msec{sec:study2}).

\begin{table}[!ht]{\small
\centering
\begin{tabular}{c|l|l}
Attack/Defense & Parameter & Setting \\
\hline
\hline
\multirow{3}{*}{\system} & perturbation threshold & $\epsilon = 8/255$\\
& learning rate & $\alpha = 1/255$\\
& maximum iterations & $n_\mathrm{iter}$ = 10\\
\hline

\multirow{3}{*}{\pgd} & PGD & $\epsilon = 8/255$\\
& learning rate & $\alpha = 1/255$\\
& maximum iterations & $n_\mathrm{iter}$ = 10\\
\hline

\multirow{2}{*}{Manifold Transformation} & network structure & [3,`average',3]\\
& random noise std & $v_{\text{noise}}$ = 0.1\\
\hline

\multirow{3}{*}{Adversarial Re-Training} & optimizer & SGD\\
 & hop steps  & $m$ = 4\\ 
& learning rate & $\alpha$ = 0.01\\
& learning rate decay & $\gamma$ = 0.1/50 epochs\\
\hline

\multirow{3}{*}{\tnet} & neuron number & $n_{\text{neuron}}$ = 2\\
& threshold & 5\\
& target value & 10 \\
\hline

\multirow{1}{*}{\strip} & number of tests & $n_{\text{test}}$ = 8\\

\end{tabular}
\caption{Default Parameter Setting \label{tab:param}}}
\end{table}

\subsubsection*{\bf B2. Curvature Profile} 

Exactly computing the eigenvalues of the Hessian matrix $\ssub{H}{x} = \sboth{\nabla}{x}{2} \ell(x)$ is prohibitively expensive for high-dimensional data. We use a finite difference approximation in our implementation. For any given vector $z$, the Hessian-vector product $\ssub{H}{x} z$ can be approximated by: 
\begin{align}
\ssub{H}{x} z = \lim_{\Delta \rightarrow 0} \frac{\ssub{\nabla}{x}\ell(x + \Delta z)-\ssub{\nabla}{x}\ell(x) }{\Delta}
\end{align}
By properly setting $\Delta$, this approximation allows us to measure the variation of gradient in $x$'s vicinity, rather than an infinitesimal point-wise curvature\cite{Dezfooli:2018:arxiv}. In practice we set $z$ as the gradient sign direction to capture the most variation:
\begin{align}
    z = \frac{\sign(\nabla \ell(x))}{\|\sign(\nabla \ell(x)) \|}
\end{align}
and estimate the magnitude of curvature as 
\begin{align}
\| \ssub{\nabla}{x}\ell(x + \Delta z)-\ssub{\nabla}{x}\ell(x)\|^2
\label{eq:approxcurv}
\end{align} 
We use \meq{eq:approxcurv} throughout the evaluation to compute the curvature profiles of given inputs.

\subsection*{\bf C. Additional Experiments}
Here we provide experiment results additional to \msec{sec:study2} and \msec{sec:study3}.

\subsubsection*{\bf C1. Detection of \tnet and \ntnet by \abs}

In addition to \strip and \nc, here we also evaluate \tnet and \ntnet against {\abs}\mcite{abs}, another state-of-the-art backdoor detection method. As the optimization of \ntnet requires white-box access to \abs, we re-implement \abs according to\mcite{abs}\footnote{\another{The re-implementation may have differences from the original \abs (\url{https://github.com/naiyeleo/ABS}).}}. In the evaluation, we set the number of seed images as 5 per class and the maximum trojan size as 400.

Similar to \nc, \abs attempts to detect potential backdoors embedded in given \dnns during model inspection. In a nutshell, its execution consists of two steps: (i) inspecting the given \dnn to sift out abnormal neurons with large elevation difference (i.e., highly active only with respect to one particular class), and (ii) identifying potential trigger patterns by maximizing abnormal neuron activation while preserving normal neuron behaviors.

To optimize the evasiveness of \ntnet with respect to \abs, we integrate the cost function (Algorithm\,2 in\mcite{abs}) into the loss terms $\lf$ and $\ls$ in Algorithm\mref{alg:watermark} and optimize the trigger $r$ and model $\theta$ respectively to minimize this cost function.

\begin{figure}[ht!]
  \centering
  \epsfig{file = 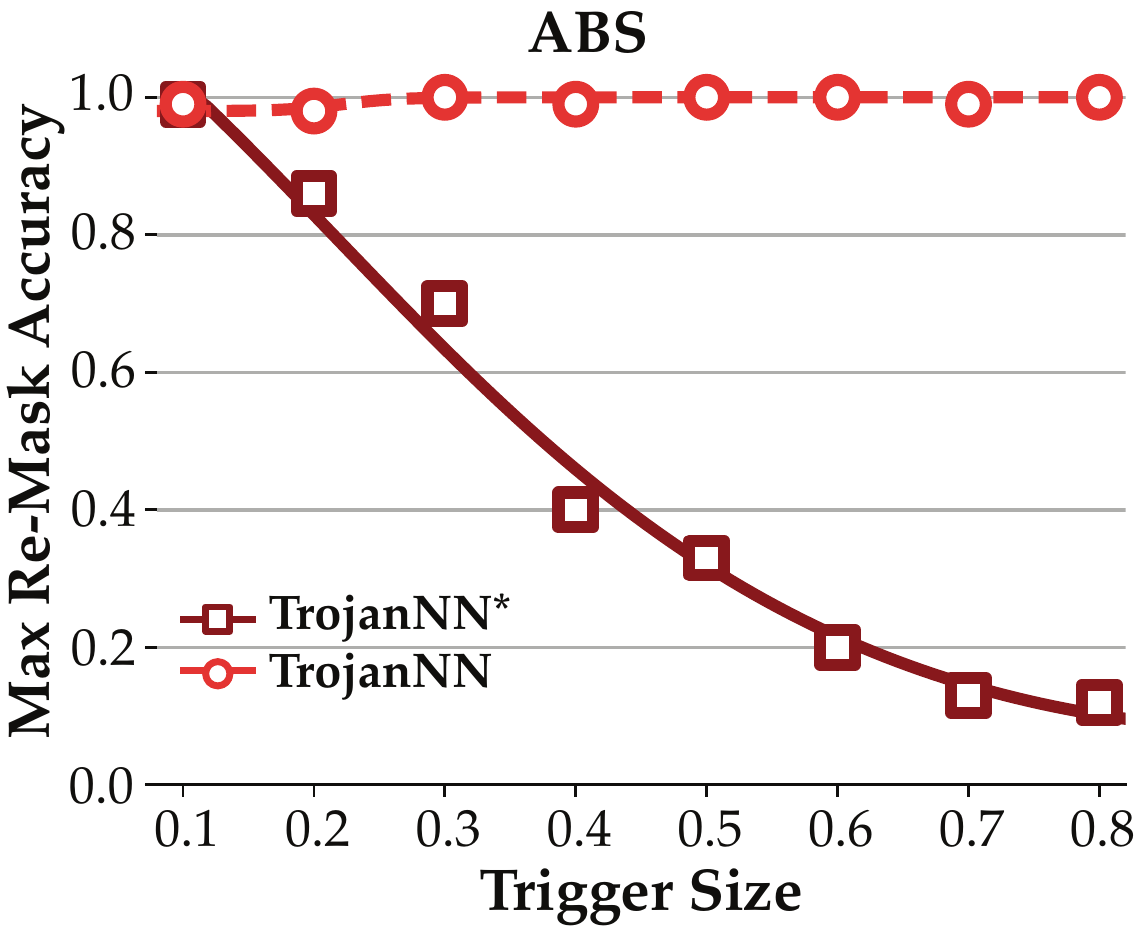, width = 50mm}
  \caption{Detection of \tnet and \ntnet by \abs on CIFAR10.\label{fig:abs}}
\end{figure}

The detection of \tnet and \ntnet by \abs on CIFAR10 is shown in Figure\mref{fig:abs}. Observe that \abs detects \tnet with close to 100\% accuracy, which is consistent with the findings in\mcite{abs}. In comparison, \ntnet is able to effectively evade \abs especially when the trigger size is sufficiently large. For instance, the detection rate (measured by maximum re-mask accuracy) drops to 40\% as the trigger size increases to 0.4. This could be explained by that larger trigger size entails more operation space for \ntnet to optimize the trigger to evade \abs.

\subsubsection*{\bf \ting{C2. Basic and Ensemble \strip against \tnet}}

\begin{figure}[ht!]
  \centering
  \epsfig{file = 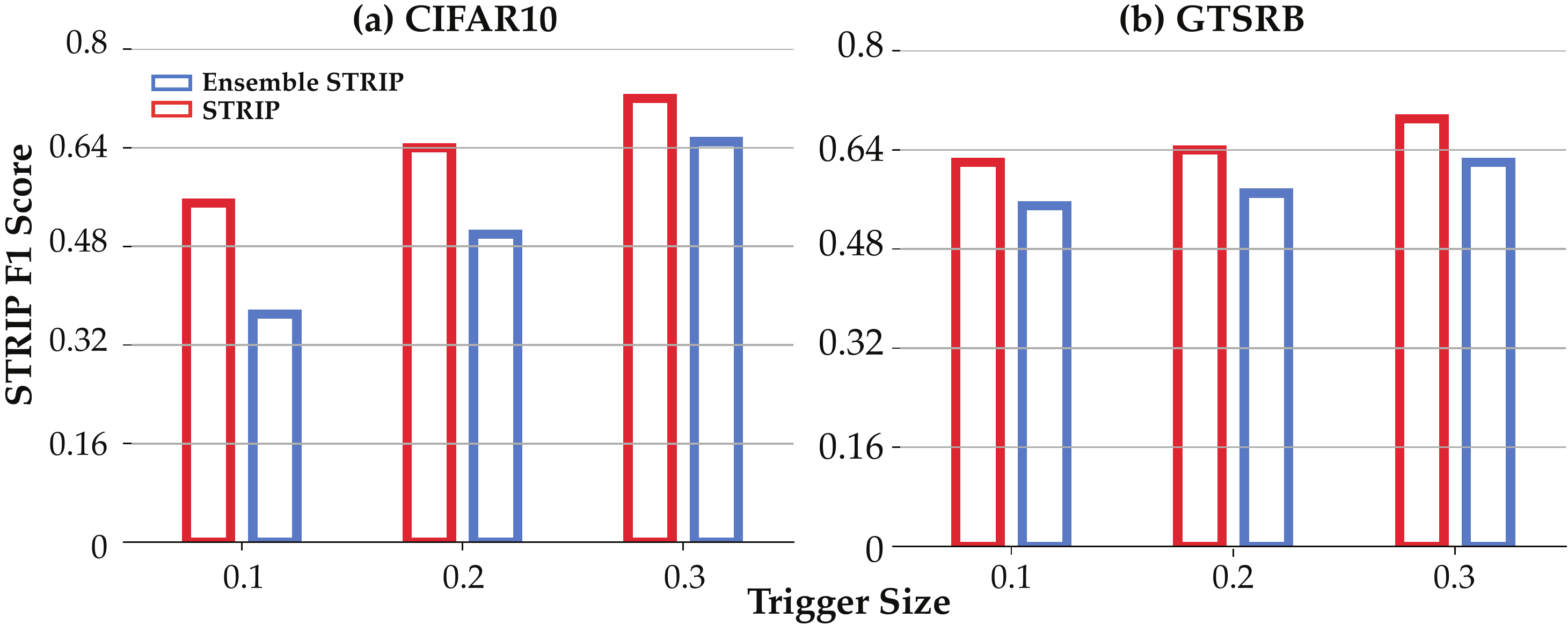, width = 85mm}
  \caption{Detection of basic and ensemble \strip against \tnet on \cifar and \gtsrb.\label{fig:detect_tnet}}
\end{figure}

\ting{Figure\mref{fig:detect_tnet} below compares the performance of basic and ensemble \strip in detecting \tnet. Interestingly, in contrary to the detection of \ntnet (Figure\mref{fig:detect_new_bar}), here the basic \strip outperforms the ensemble version. This may be explained as follows. As \ntnet is optimized to evade both \strip and \nc, to effectively detect it, ensemble \strip needs to balance the metrics of both detectors; in contrast, \tnet is not optimized with respect to either detector. The compromise of ensemble \strip results in its inferior performance compared with the basic detector.}

\subsection*{\bf D. Symbols and Notations}
Table\mref{tab:symbol} summarizes the important notations in the paper.

\begin{table}[!ht]{\small
  \centering
\begin{tabular}{r|l}
  Notation & Definition\\
  \hline
  \hline
$\bx, \ax$ & benign, adversarial inputs\\
$\btheta, \atheta$ & benign, poisoned \dnns\\
$\ay$ & adversary's target class \\
$\kappa$ & misclassification confidence threshold\\
$\mathcal{D}, \mathcal{R}, \mathcal{T}$ & training, reference, target sets\\
$\ell, \ls, \lf$ & attack efficacy, specificity, fidelity losses\\
$\phi$ & leverage effect coefficient\\
$\alpha$ & learning rate\\
$\epsilon, \delta$ & thresholds of input, model perturbation \\
\end{tabular}
\caption{Symbols and notations. \label{tab:symbol}}}
\end{table}